\documentclass[journal]{IEEEtran}
\ifCLASSINFOpdf
\else
\fi
\usepackage{graphicx}
\usepackage{epstopdf}
\usepackage{epsfig,amsmath,amssymb,amsthm,array}
\usepackage[square]{natbib}

\newcommand{\ed}{\stackrel{\triangle}{=}}

\newcommand{\ben}{\begin{equation*}}
\newcommand{\een}{\end{equation*}}
\newcommand{\bl}{\begin{align}}
\newcommand{\el}{\end{align}}
\newcommand{\bln}{\begin{align*}}
\newcommand{\eln}{\end{align*}}
\newcommand{\bld}{\begin{aligned}}
\newcommand{\eld}{\end{aligned}}
\newcommand{\be}{\begin{equation}}
\newcommand{\ee}{\end{equation}}
\newcommand{\ba}{\begin{eqnarray}}
\newcommand{\ea}{\end{eqnarray}}
\newcommand{\ban}{\begin{eqnarray*}}
\newcommand{\ean}{\end{eqnarray*}}

\newcommand{\bit}{\begin{itemize}}
\newcommand{\eit}{\end{itemize}}

\newtheorem{theorem}{Theorem}

\begin{document}
%
% paper title
% Titles are generally capitalized except for words such as a, an, and, as,
% at, but, by, for, in, nor, of, on, or, the, to and up, which are usually
% not capitalized unless they are the first or last word of the title.
% Linebreaks \\ can be used within to get better formatting as desired.
% Do not put math or special symbols in the title.
\title{ Ranking and Selection as Stochastic Control}
%
%
% author names and IEEE memberships
% note positions of commas and nonbreaking spaces ( ~ ) LaTeX will not break
% a structure at a ~ so this keeps an author's name from being broken across
% two lines.
% use \thanks{} to gain access to the first footnote area
% a separate \thanks must be used for each paragraph as LaTeX2e's \thanks
% was not built to handle multiple paragraphs
%

\author{ Yijie~Peng,~\IEEEmembership{}
        Edwin K. P.~Chong,~\IEEEmembership{} Chun-Hung~Chen,~\IEEEmembership{} and~ Michael C.~Fu~\IEEEmembership{}
       % <-this % stops a space
\thanks{Yijie Peng is with the Department
of Industrial Engineering and Management, Peking University, 
Beijing, 100871 China e-mail:  pengy10@fudan.edu.cn. He is the corresponding author. }   

\thanks{Edwin Chong is with the Department
of Electrical and Computer Engineering, Colorado State University, Fort Collins,
CO, 80523-1373
 USA e-mail: edwin.chong@colostate.edu.}    
       
\thanks{Chun-Hung~Chen is with the Department
of System Engineering and Operations Research, George Mason University, Fairfax,
VA, 22030 USA e-mail: cchen9@gmu.edu.}% <-this % stops a space
% <-this % stops a space

\thanks{Michael Fu is with the R.H. Smith School of Business and Institute for Systems Research, University of Maryland, College Park,
MD, 20742 USA e-mail: mfu@isr.umd.edu.}

%\thanks{Manuscript received April 19, 2005; revised August 26, 2015.}
}

\maketitle

% As a general rule, do not put math, special symbols or citations
% in the abstract or keywords.
\begin{abstract}
Under a Bayesian framework, we formulate the fully sequential sampling and selection decision in statistical ranking and selection as a stochastic control problem, and derive the associated Bellman equation. Using value function approximation,  we derive an approximately optimal allocation policy. We show that this policy is not only computationally efficient but also possesses both one-step-ahead and asymptotic optimality for independent normal sampling distributions. Moreover, the proposed allocation policy is easily generalizable  in the approximate dynamic programming paradigm.
\end{abstract}

% Note that keywords are not normally used for peerreview papers.
\begin{IEEEkeywords}
simulation, ranking and selection, stochastic control, Bayesian, dynamic sampling and selection.
\end{IEEEkeywords}

% For peer review papers, you can put extra information on the cover
% page as needed:
% \ifCLASSOPTIONpeerreview
% \begin{center} \bfseries EDICS Category: 3-BBND \end{center}
% \fi
%
% For peerreview papers, this IEEEtran command inserts a page break and
% creates the second title. It will be ignored for other modes.
\IEEEpeerreviewmaketitle

\section{Introduction}
In this paper, we consider a simulation optimization problem of choosing the highest mean alternative from a finite set of alternatives, where the means are unknown and must be estimated by statistical sampling. In simulation, this problem is often called statistical ranking and selection (R\&S) problem (see \cite{BSG95}). Applications of R\&S include selecting the best alternative from many complex discrete event dynamic systems (DEDS) that are computationally intensive to simulate (see \cite{qing2013stochastic}), and finding the most effective drug from different alternatives, where the economic cost of each sample for testing the effectiveness of the drug is expensive (see \cite{powell2012ranking}). 
Broadly speaking, there are two main approaches in R\&S (\cite{goldsman1998comparing} and \cite{chen2011stochastic}). The first approach allocates samples to guarantee the probability of correct selection (PCS) up to a pre-specified level (\cite{rinott1978two}, \cite{kim2006selecting}, \cite{kim2013statistic}), whereas the second approach maximizes the PCS (or other similar metric) subject to a given sampling budget (\cite{chen2000simulation}, \cite{chick2001new},  \cite{lee2012approximate},  \cite{pasupathy2014stochastically}). 

The earliest sampling allocation schemes use two-stage procedures (e.g., \cite{rinott1978two}, \cite{chen2000simulation}, \cite{chick2001new}), where unknown parameters are estimated in the first stage. More recently, fully sequential sampling allocation procedures have been developed (\cite{kim2001fully}, \cite{hong2005tradeoff}, \cite{frazier2014fully}).  
 In the Bayesian framework, \cite{chen2006efficient}, \cite{frazier2008knowledge}, and \cite{chick2010sequential} proposed sequential algorithms  by allocating each replication to maximize the posterior information gains one step ahead; \cite{chick2009economic} and \cite{chick2012sequential} provided sequential policies analogous to a multi-armed bandit problem and used a continuous-time approximation to solve their Bellman equation; \cite{peng2014dynamic} offered a sequential rule achieving the asymptotically optimal sampling rate of the PCS; and \cite{peng2015asymptotic} developed a sequential algorithm that possesses both one-step-ahead and asymptotic optimality.

Previous work using the Bayesian framework approached the difficult dynamic R\&S problem by replacing the sequential sampling and selection  decisions  with a more tractable surrogate optimization problem.
In this work, we formulate the R\&S problem as a stochastic control problem (SCP) and derive  the associated Bellman optimality equation, which requires care due to the interaction between the sampling allocation policy and the posterior distribution.
 We show that under a canonical condition in R\&S, the sampling allocation decision does not affect the Bayesian posterior distributions conditional on the information of sample observations; thus the SCP is proved to be a Markov decision process (MDP). To the best of our knowledge, this is the first work to study
 R\&S as an SCP and use MDP to analyze it.

We then analyze the optimal allocation and selection (A\&S) policy of the resulting MDP and prove that a commonly used selection policy of selecting the alternative with the largest sample mean is asymptotically consistent with the optimal selection policy under some mild conditions. The size of the state space of the sampling allocation policy for independent discrete sampling distributions is shown to only grow polynomially with respect to the number of allocated replications if the numbers of alternatives and the possible outcomes of the discrete distributions are fixed, but will have an exponential growth rate when the number of alternatives and the number of possible outcomes of the discrete distributions grow together. 

Sampling from independent normal distributions is a standard assumption in the R\&S literature, so we focus on this setting. 
In contrast to the usual approach of replacing the SCP with a tractable approximate surrogate (static) optimization problem,
we address the SCP directly by approximating the value function, as in approximate dynamic programming (ADP)  (see \cite{powell2007approximate}).
The value function approximation (VFA) using a simple feature of the value function yields an approximately optimal allocation policy (AOAP) that  is not only computationally efficient,  but also possesses both one-step-ahead and asymptotic optimality. In addition, the VFA approach is easily generalizable in the ADP paradigm. For example, we show how to extend the AOAP to a multi-step look-ahead sampling allocation procedure, and how to obtain an efficient sampling algorithm for a low-confidence scenario (see \cite{peng2016gradient}) by implementing an off-line learning algorithm. 

 The rest of the paper is organized as follows: Section \ref{pfl} formulates the SCP in R\&S, and the associated Bellman equation is derived in Section \ref{sc}. Section \ref{asp} offers further analysis on the optimal A\&S policy, and Section \ref{asn} focuses on the approximations of the optimal A\&S policy for normal sampling distributions. Numerical results are given in Section~\ref{nex}. The last section offers conclusions.
 \section{Problem Formulation}\label{pfl}
 Among $k$ alternatives with unknown means $\mu_i$, $i=1,\ldots,k$, our objective is to find the best alternative defined by $$\langle 1 \rangle\ed\arg\max_{i=1,\ldots,k}\mu_i,$$ where each $\mu_i$ is estimated by sampling.  
 Let  $X_{i,t}$ be the $t$-th replication for alternative $i$. Suppose $X_{t}\ed(X_{1,t},\ldots,X_{k,t})$, $t\in \mathbb{Z}^{+}$, follows an independent and identically distributed (i.i.d.)\ joint sampling distribution, i.e., $X_{t}\sim Q(\cdot;\theta)$, with a density (or probability mass function) $q(\cdot;\theta)$,  where $\theta\in\Theta$ comprises all unknown parameters in the parametric family. The marginal distribution of alternative $i$ is denoted by  $Q_i(\cdot;\theta_i)$, with a density $q_i(\cdot;\theta_i)$, where $\theta_i$ comprises all unknown parameters in the marginal distribution. Generally, $\mu_i\in  \theta_i$, $i=1,\ldots,k$, and $(\theta_1,\ldots,\theta_k)\in \theta$. 
 In addition, we assume the  unknown parameter follows a prior distribution, i.e., $\theta\sim F(\cdot;\zeta_{0})$, where $\zeta_0$ contains all hyper-parameters for the parametric family of the prior distribution. 
  
  We define the two parts of an A\&S policy. The allocation policy is a sequence of mappings $\mathcal{A}_t(\cdot)=(A_1(\cdot),\ldots,A_t(\cdot))$, where 
  $A_{t}(\mathcal{E}_{t-1}^a)\in \{1,\ldots,k\}$, which allocates the $t$-th sample to  an alternative based on information set $\mathcal{E}_{t-1}^a$ collected through all previous steps. The information set at step $t$ is given by  
  $$\mathcal{E}_t^a\ed\left\{\mathcal{A}_t(\mathcal{E}_{t-1}^a); \mathcal{E}_t\right\},$$
  where  $\mathcal{E}_t$ contains all sample information and prior information $\zeta_0$. 
 Define $A_{i,t}(\mathcal{E}_{t-1}^a)\ed{\bf1}\left\{A_{t}(\mathcal{E}_{t-1}^a)=i\right\}$.  
 The information collection procedure following a sampling allocation policy in R\&S problem is illustrated in Figure \ref{fig.1} for allocating four samples among three alternatives. Given prior information $\mathcal{E}_0$, collected information set $\mathcal{E}_4^a$ is determined by the two tables in the figure. The allocation decision represented by the table at the bottom determines the (bold) observable elements  in the table on the top. 
   \begin{figure}[tb]
    \centering
   \begin{tabular}{c c c }
   $X_{1,1}$  &${\bf X_{2,1}}$ &$X_{3,1}$\\
   ${\bf X_{1,2}}$  & $X_{2,2}$&$X_{3,2}$\\
   ${\bf X_{1,3}}$  & $X_{2,3}$&$X_{3,3}$\\
   $X_{1,4}$  & $X_{2,4}$&${\bf X_{3,4}}$
   \end{tabular}\vspace*{3mm}
      \begin{tabular}{c c c }
      $A_{1,1}(\zeta_0)=0$  &$A_{2,1}(\zeta_0)=1$ &$A_{3,1}(\zeta_0)=0$\\
      $A_{1,2}(\mathcal{E}_1^a)=1$  & $A_{2,2}(\mathcal{E}_1^a)=0$&$A_{3,2}(\mathcal{E}_1^a)=0$\\
      $A_{1,3}(\mathcal{E}_2^a)=1$  & $A_{2,3}(\mathcal{E}_2^a)=0$&$A_{3,3}(\mathcal{E}_2^a)=0$\\
      $A_{1,4}(\mathcal{E}_3^a)=0$  & $A_{2,4}(\mathcal{E}_3^a)=0$&$A_{3,4}(\mathcal{E}_3^a)=1$
      \end{tabular}
   \vspace*{0.1in}
   \caption{ Illustration of information collection procedure in R\&S. 
   }
   \label{fig.1}\normalsize
   \end{figure}
   The sampling decision and information flow have an interactive relationship shown in Figure \ref{fig.2}. 
    \begin{figure}[tb]
    \begin{center}
    \begin{picture}(420,40)(20,0)
    \put (33, 20) {$\zeta_0$}
    \put (48, 23) {\vector(1, 0){10}} 
    \put (63, 20) {$\mathcal{E}_1^a=\left\{ A_1(\zeta_0)=2;\mathcal{E}_1\right\}$} 
    %\put (120, 3) {\vector(1, 0){10}} 
     %   \put (150, 0) {$\mathcal{E}_1^a=\left\{ A_1(\zeta_0)=2;\mathcal{E}_1\right\}$}
    \put (178, 23) {\vector(1, 0){8}}  
    \put (70, 0) {$\cdots$} 
    \put (88, 3) {\vector(1, 0){8}}\put (100, 0) {$\mathcal{E}_4^a=\left\{A_1(\zeta_0)=2,\ldots,A_4(\mathcal{E}_3^a)=3; \mathcal{E}_4 \right\}$}
    \end{picture}
    \end{center}
    \caption{Interaction between sampling allocation decision and information flow in Figure \ref{fig.1}.}
    % \hspace{1in}
    \label{fig.2}
    \end{figure}
    The sampling decision and the information set are nested in each other as $t$ evolves. We reorganize the (allocated) sample observations by putting them together and ordering them in chronological arrangement, i.e., $\bar{X}_i^{(t)}\ed(\bar{X}_{i,1},\ldots,\bar{X}_{i,t_i})$, where 
    $t_i\ed\sum_{\ell=1}^{t}A_{i,\ell}(\mathcal{E}_{\ell-1}^a)$, $i=1,\ldots,k$. Although $t_i$ is also a map from the information set, we suppress the argument for simplicity. For the example in Figure \ref{fig.1}, we specifically illustrate how to reorganize the sample observations in Figure \ref{fig.3}. 
       \begin{figure}[tb]
        \centering
       \begin{tabular}{c c c }
       $X_{1,1}$  &${\bf X_{2,1}}$ &$X_{3,1}$\\
       ${\bf X_{1,2}}$  & $X_{2,2}$&$X_{3,2}$\\
       ${\bf X_{1,3}}$  & $X_{2,3}$&$X_{3,3}$\\
       $X_{1,4}$  & $X_{2,4}$&${\bf X_{3,4}}$
       \end{tabular}\quad $\implies 
       $ 
       \vspace*{3mm}
             
             \begin{tabular}{c c c }
                 $\bar{X}_{1,1}={\bf X_{1,2}}$  &$\bar{X}_{2,1}={\bf X_{2,1}}$ &$\bar{X}_{3,1}={\bf X_{3,4}}$\\
                 $\bar{X}_{1,2}={\bf X_{1,3}}$  & &\\
                 \end{tabular}
       \vspace*{0.1in}
       \caption{ Reorganization of sample observations. 
       }
       \label{fig.3}\normalsize
       \end{figure}
       We have $\mathcal{E}_t=\{ \zeta_0,  \bar{X}_1^{(t)},\ldots,\bar{X}_k^{(t)} \}$. 
   
    The selection policy is a map $\mathcal{S}(\mathcal{E}_T^a)\in \{1,\ldots,k\}$, which makes the final selection at step $T$ and  indicates the best alternative chosen by the A\&S algorithm.
   The final reward for selecting an alternative is a function of $\theta$, given the selection decision, i.e., $V(\theta;i)|_{i=\mathcal{S}}$. In R\&S, two of the most frequently used candidates for the final reward are
   \begin{align*}
   V_{P}(\theta;i)\ed{\bf1}\{ i=\langle 1 \rangle \},\quad V_{E}(\theta;i)\ed\mu_{i}-\mu_{\langle 1 \rangle},
   \end{align*}
   where %$\langle 1 \rangle\ed\arg\max_{i=1,\ldots,k}\mu_i$, and 
   the subscripts $P$ and $E$  stand for PCS and expected opportunity cost (EOC), respectively, and $\langle i \rangle$, $i=1,\ldots,k$, are order statistics s.t. $\mu_{\langle 1 \rangle}>\cdots>\mu_{\langle k\rangle}$. If the alternative selected as the best is the true best alternative, $V_P$ is one, otherwise $V_P$ is zero; $V_E$ is the difference between the mean of the selected alternative and the mean of the true best alternative, which measures the economic opportunity cost (regret) of the selection decision. Notice that the values of the final rewards $V_P$ and $V_E$ are unknown due to the uncertainty of parameter $\theta$, which is quantified by the prior distribution of the parameter in the Bayesian framework. 
   
   We formulate the dynamic decision in R\&S by a SCP as follows. 
   Under the Bayesian framework, the expected payoff for an A\&S policy $(\mathcal{A},\mathcal{S})$, where $\mathcal{A}\ed\mathcal{A}_T$, in the SCP can be defined recursively by 
   \begin{align*}\label{s1} &V_T(\mathcal{E}_T^a;\mathcal{A},\mathcal{S})\ed\left.\mathbb{E}\left[ V(\theta;i)|\mathcal{E}_T^a\right]\right|_{i=\mathcal{S}(\mathcal{E}_T^a)}\\
   =&\left.\int_{\theta\in \Theta} V(\theta;i)~ F(d\theta|\mathcal{E}_T^a)\right|_{i=\mathcal{S}(\mathcal{E}_T^a)},
   \end{align*}
  where  $F(\cdot|\mathcal{E}_t^a)$ is the posterior distribution of $\theta$ conditioned on the information set $\mathcal{E}_t^a$, and  $d\cdot$ in $d\theta$ stands for Lebesgue measure for continuous distributions and the counting measure for discrete distributions,
  \begin{equation*}
    \begin{aligned}\label{s2} V_{t}&(\mathcal{E}_{t}^a;\mathcal{A},\mathcal{S})\ed\left.\mathbb{E}\left[ V_{t+1}(\mathcal{E}_{t}^a\cup\{X_{i,t+1}\};\mathcal{A},\mathcal{S})|\mathcal{E}_{t}^a\right]\right|_{i=A_{t+1}(\mathcal{E}_{t}^a)}\\
    &=\left.\int_{\mathcal{X}_i}V_{t+1}(\mathcal{E}_{t}^a\cup\{x_{i,t+1}\};\mathcal{A},\mathcal{S})~ Q_i(d x_{i,t+1}|\mathcal{E}_{t}^a)\right|_{i=A_{t+1}(\mathcal{E}_{t}^a)},
    \end{aligned} 
    \end{equation*}
 where $\mathcal{X}_i$ is the support of $X_{i,t+1}$, and $Q_i(\cdot|\mathcal{E}_{t}^a)$  is the predictive distribution for $X_{i,t+1}$ conditioned on the information set $\mathcal{E}_{t}^a$. The posterior and predictive distributions can be calculated using Bayes rule:
 \begin{align}\label{pd1}
 F(d\theta|\mathcal{E}_t^a)=\frac{L(\mathcal{E}_t^a;\theta)~ F(d\theta;\zeta_0)}{\int_{\theta\in\Theta}L(\mathcal{E}_t^a;\theta) ~F(d\theta;\zeta_0)},
 \end{align}
 and
 \begin{align}\label{prd1}
 Q_i(d x_{i,t+1}|\mathcal{E}_t^a)=\frac{\int_{\theta\in\Theta} Q_i(d x_{i,t+1};\theta_i)~L(\mathcal{E}_t^a;\theta)~ F(d\theta;\zeta_0)}{\int_{\theta\in\Theta}L(\mathcal{E}_t^a;\theta) ~F(d\theta;\zeta_0)},
 \end{align}
 where $L(\cdot)$ is the likelihood of the samples. The  posterior and predictive distributions for specific sampling distributions will be discussed in the next section. 
 With the formulation of the SCP, we define an optimal A\&S policy as  
 \begin{equation}\label{sp}
(\mathcal{A}^{*},\mathcal{S}^{*})\ed\sup_{\mathcal{A},\mathcal{S}}V_0(\zeta_0;\mathcal{A},\mathcal{S})~.
 \end{equation} 
\section{ R\&S as Stochastic Control}\label{sc}
In Section \ref{as}, we establish the Bellman equation for SCP (\ref{sp}). In Section \ref{cj}, we show that the information set determining the posterior and predictive distributions can be further reduced to hyper-parameters by using conjugate priors.
 \subsection{Optimal A\&S Policy}\label{as}
 To avoid having to keep track of the entire sampling allocation policy history, the following theorem establishes that the posterior and predictive distributions at step $t$ are determined by 
 $\mathcal{E}_t$; thus, if we define  $\mathcal{E}_t$ as the state at step $t$, then SCP (\ref{sp}) satisfies the optimality equation of an MDP. 
 \begin{theorem}\label{thm1} Under the Bayesian framework introduced in Section \ref{pfl}, 
  the posterior distribution (\ref{pd1}) of $\theta$ conditioned on $\mathcal{E}_T$ and the predictive distribution (\ref{prd1}) of $X_{i,t+1}$ conditioned on $\mathcal{E}_t$  are independent of the allocation policy $\mathcal{A}$. 
 \end{theorem}
 \begin{IEEEproof}
 At any step $t$, all replications except for the replication of the alternative being sampled, $i=A_t(\mathcal{E}_{t-1}^a)$, are missing.  The likelihood of observations collected by the sequential sampling procedure though $t$ steps is given by 
 \begin{equation}\label{smar}
 \begin{aligned}
L&(\mathcal{E}_t^a;\theta)=\int\cdots\int_{\mathcal{X}^t}\prod_{\ell=1}^{t} q(x_{\ell};\theta)\prod_{i=1}^{k}\left\{ A_{i,\ell}(\mathcal{E}_{\ell-1}^a)~\delta_{X_{i,\ell}} (d x_{i,\ell})\right.\\&\left.\qquad\qquad\qquad\qquad\qquad+(1-A_{i,\ell}(\mathcal{E}_{\ell-1}^a))~d x_{i,\ell}\right\}\\
&=\left(\sum_{i=1}^{k} A_{i,t} (\mathcal{E}_{t-1}^{a})~ q_i(X_{i,t};\theta_i)\right) \int\cdots\int_{\mathcal{X}^{t-1}}\prod_{\ell=1}^{t-1} q(x_{\ell};\theta)\\
&\times\prod_{i=1}^{k}\left\{ A_{i,\ell}(\mathcal{E}_{\ell-1}^a)~\delta_{X_{i,\ell}} (d x_{i,\ell})+(1-A_{i,\ell}(\mathcal{E}_{\ell-1}^a))~d x_{i,\ell}\right\}\\
 &=\prod_{\ell=1}^{t}\left(\sum_{i=1}^{k} A_{i,\ell} (\mathcal{E}_{\ell-1}^{a})~ q_i(X_{i,\ell};\theta_i)\right)=\prod_{i=1}^{k}\prod_{\ell=1}^{t_i} q_i(\bar{X}_{i,\ell};\theta_i),
 \end{aligned}
 \end{equation}
 where $\mathcal{X}\ed\mathcal{X}_1\times \cdots\times\mathcal{X}_k$ and $\delta_{x}(\cdot)$ is the delta-measure with a mass point at $x$. 
  The first equality in (\ref{smar}) holds because $X_{\ell}$, $\ell\in\mathbb{Z}^{+}$, are assumed to be i.i.d. and  the $t$-th replication $X_t$  is independent of the information flow before $t$ step, i.e., $\zeta_0\cup\{\mathcal{E}_{\ell}^a\}_{\ell=1}^{t-1}$, by  construction of the information set; thus the variables of the missing replications at step $t$ in the joint density are integrated out, leaving only the marginal likelihood of the observation at step $t$.  By using the same argument inductively, the second equality in (\ref{smar}) holds. The last equality in (\ref{smar}) holds because the product operation is commutative. 
With (\ref{smar}), we can denote the likelihood  as $L(\mathcal{E}_t;\theta)$, since the information set $\mathcal{E}_t$ completely determines the likelihood. 
  
  Following Bayes rule, the posterior distribution of $\theta$ is
 \begin{equation}\label{pd}
 \begin{aligned} F(d\theta|\mathcal{E}_T^a)
 &=\frac{ L(\mathcal{E}_T;\theta) ~F(d\theta;\zeta_0)}{\int_{\theta\in\Theta} L(\mathcal{E}_T;\theta) ~F(d\theta;\zeta_0) }
 \\&= \frac{\prod_{i=1}^{k}\prod_{t=1}^{T_i} q_i(\bar{X}_{i,t};\theta_i)~ F\left(d\theta;\zeta_0\right)}{\int_{\theta\in\Theta} \prod_{i=1}^{k}\prod_{t=1}^{T_i} q_i(\bar{X}_{i,t};\theta_i)~ F\left(d\theta;\zeta_{0}\right)},\end{aligned}
 \end{equation}
 which is independent of the allocation policy $\mathcal{A}$, conditioned on $\mathcal{E}_T$. With (\ref{pd}), we can denote the posterior distribution as $F(d\theta|\mathcal{E}_T)$, since the information set $\mathcal{E}_T$ completely determines the posterior distribution. 
 Similarly, the predictive distribution of $X_{i,t+1}$ is 
  \begin{equation}\label{prd}
  \begin{aligned}Q_i&(dx_{i,t+1}|\mathcal{E}_{t}^a)
  =\frac{\int_{\theta\in\Theta}Q_i(dx_{i,t+1};\theta_i)~L(\mathcal{E}_t;\theta)~F(d\theta;\zeta_0)}{\int_{\theta\in\Theta} L(\mathcal{E}_t;\theta)~F(d\theta;\zeta_0)} \\
  =&\frac{\int_{\theta\in\Theta}Q_i(dx_{i,t+1};\theta_i)~\prod_{i=1}^{k}\prod_{\ell=1}^{t_i} q_i(\bar{X}_{i,\ell};\theta_i)~ F\left(d\theta;\zeta_0\right)}{\int_{\theta\in\Theta}  \prod_{i=1}^{k}\prod_{\ell=1}^{t_i} q_i(\bar{X}_{i,\ell};\theta_i)~ F\left(d\theta;\zeta_0\right)},\end{aligned}
  \end{equation}
 which is independent of the allocation policy $\mathcal{A}$, conditioned on $\mathcal{E}_t$. With (\ref{prd}), we can denote the  predictive distribution of $X_{i,t+1}$ as $Q_i(dx_{i,t+1}|\mathcal{E}_{t})$, since  $\mathcal{E}_t$ completely determines the predictive distribution.
\end{IEEEproof}
 \noindent\textbf{Remark.} 
 The interaction between sampling allocation policy and posterior distribution has also been studied by  \cite{gorder2014ranking}, but they introduced a monotone missing pattern that is not satisfied by the sequential sampling mechanism assumed in our paper. 
 If the sampling distribution is assumed to be independent, i.e., $Q(x_t;\theta)=\prod_{i=1}^{k} Q_i(x_{i,t};\theta)$, the missing pattern can be fitted into a missing at random (MAR) paradigm studied in incomplete data analysis.  MAR means that the missing rule is independent of the missing data, given the observations; see Chapter 2 of \cite{kim2013statistical} for a rigorous definition. If the sampling distribution is dependent, the sequential information collection procedure in our work does not satisfy the classic MAR paradigm. For the example in Figure \ref{fig.1}, we can see that if the sampling distribution is not independent, the missing rule, say $A_{1,4}(\mathcal{E}_3^a)$, could be dependent on the missing data, say $X_{3,1}$, since $X_{3,1}$ and $X_{2,1}$ are dependent  and $X_{2,1}\in \mathcal{E}_3^a$. Even without the MAR condition, we can still prove our conclusion because of two facts: (1) the replications, i.e., $X_t$, $t\in\mathbb{Z}^{+}$, of the sampling distribution are assumed to be independent; (2) the allocation decision at step $t$, i.e.,  $A_t(\mathcal{E}_{t-1}^a)$, only depends on the information set collected at the step $t-1$ in our setting. We call the special structure of sequential sampling decision in R\&S  sequentially MAR (SMAR). 
 
 Dependence in the sampling distribution is often introduced by using  common random numbers (CRN) to enhance the efficiency of R\&S (see \cite{fu2007simulation} and \cite{peng2012efficient}). Although dependence in the sampling distribution is not a problem, our proof for Theorem \ref{thm1} does not apply if there is dependence between replications, because $X_{i,t}$ and $A_{j,t}(\mathcal{E}_{t-1}^a)$, $j=1,\ldots,k$, could be dependent in this case. The i.i.d. assumption for replications, assumed in our paper, is a canonical condition in R\&S. \\ 
 
 \noindent\underline{Bellman Equation:}\\
 
  With the conclusion of Theorem \ref{thm1}, the R\&S problem is an MDP with state $\mathcal{E}_t$, action $A_{t+1}$ for $0\leq t<T$ and $\mathcal{S}$ for $t=T$, no reward for $0\leq t<T$ and $V_T(\mathcal{E}_T;\mathcal{S})$ for $t=T$, and the following transition for $0\leq t<T$:
  \begin{align*}&\{ \zeta_0, \bar{X}_1^{(t)},\ldots,\bar{X}_k^{(t)} \}
  \\
  &\quad\to \{ \zeta_0,  \bar{X}_1^{(t)},\ldots,\bar{X}_{i}^{(t)},X_{i,t+1},\ldots,\bar{X}_k^{(t)} \}|_{i=A_{t+1}},\end{align*}
  where $X_{i,t+1}\sim Q_{i}(\cdot|\mathcal{E}_t)$, $i=A_{t+1}$.
  Then, we can recursively compute the optimal A\&S policy $(\mathcal{A}^{*},\mathcal{S}^{*})$ of the SCP (\ref{sp}) by the following Bellman equation:
 \begin{equation}\label{be1}V_T(\mathcal{E}_T)\ed\left.V_T(\mathcal{E}_T; i)\right|_{i=\mathcal{S}^{*}(\mathcal{E}_T)},\end{equation}
  where $V_T(\mathcal{E}_T; i)\ed\mathbb{E}\left[ V(\theta;i)|\mathcal{E}_T \right]$, and 
   \begin{equation*}\label{osp}\mathcal{S}^{*}(\mathcal{E}_T)=\arg\max_{i=1,\ldots,k}V_T(\mathcal{E}_T; i),\end{equation*}
   and for $0\leq t<T$,
   \begin{equation}\label{be2}V_{t}(\mathcal{E}_{t})\ed\left.V_{t}(\mathcal{E}_{t};i )\right|_{i=A_{t+1}^{*}(\mathcal{E}_t)},\end{equation}
where $V_{t}(\mathcal{E}_{t};i)\ed\mathbb{E}\left[ V_{t+1}(\mathcal{E}_{t}, X_{i,t+1}) |\mathcal{E}_{t}\right]$, and 
\begin{equation*}\label{oap}A_{t+1}^{*}(\mathcal{E}_{t})=\arg\max_{i=1,\ldots,k}V_{t}(\mathcal{E}_{t};i)~.\end{equation*}
For an MDP, the equivalence between the optimal policy of the SCP, i.e., (\ref{sp}), and the optimal policy determined by the Bellman equation, i.e., (\ref{be1}) and (\ref{be2}), can be established straightforwardly by induction. The equivalence discussion can be found in Proposition 1.3.1 of \cite{bertsekas1995dynamic}.  
 \subsection{Conjugacy}\label{cj}
 Notice that the dimension of the state space of the MDP in the last section grows as the step grows. Under the conjugate prior, the  information set $\mathcal{E}_t$ can be completely determined by the posterior hyper-parameters, i.e.,
 $\mathcal{E}_t=\zeta_t$. Thus, the dimension of the state space is the dimension of the hyper-parameters, which is fixed at any step. 
 We provide specific forms for the conjugacy of independent Bernoulli distributions and independent normal distributions with known variances. \\

\noindent \underline{1. Bernoulli Distribution}\\

The Bernoulli distribution is a discrete distribution with probability mass function (p.m.f.): $q_i(1;\theta_i)=\theta_i$ and $q_i(0;\theta_i)=1-\theta_i$, so the mean of alternative $i$ is $\mu_i=\theta_i$. The conjugate prior for the Bernoulli distribution is a beta distribution with density $f_i(\theta_i;\alpha_i^{(0)},\beta_i^{(0)})$, where 
$$f_i(\theta_i;\alpha_i,\beta_i)=\frac{\theta_i^{\alpha_i-1}\left(1-\theta_i\right)^{\beta_i-1}}{\int_{0}^{1}\theta_i^{\alpha_i-1}\left(1-\theta_i\right)^{\beta_i-1}d\theta_i},~\theta_i\in[0,1],~\alpha_i,\beta_i>0.$$
With (\ref{pd}) and (\ref{prd}),  the posterior distribution of $\theta_i$ is
$$F_i(d\theta_i;\zeta_{t,i})=f_i(\theta_i;\alpha_i^{(t)},\beta_i^{(t)}) d\theta_i,$$
where $\zeta_{t,i}\ed(\alpha_i^{(t)},\beta_i^{(t)})$, and 
\begin{align*}
&\alpha_i^{(t)}=\alpha_i^{(0)}+t_i m_i^{(t)},\quad 
\beta_i^{(t)}=\beta_i^{(0)}+t_i(1-m_i^{(t)}),\\& m_i^{(t)}\ed\frac{\sum_{\ell=1}^{t_i}  \bar{X}_{i,\ell}}{t_i},\end{align*}
and the predictive p.m.f. of $X_{i,t+1}$ is
$$q_i(1;\zeta_{t,i})=\gamma_i^{(t)},\quad q_i(0;\zeta_{t,i})=1-\gamma_i^{(t)},$$
where 
$$\gamma_i^{(t)}\ed\frac{\alpha_i^{(t)}}{\alpha_i^{(t)}+\beta_i^{(t)}}~.$$
Assuming $\gamma_i^{(0)}=\gamma_j^{(0)}$, if $t_i=t_j$ and $m_i^{(t)}>m_j^{(t)}$, then $\gamma_i^{(t)}>\gamma_j^{(t)}$, and if
    $m_i^{(t)}=m_j^{(t)}$ and $t_i>t_j$, then $\gamma_i^{(t)}>\gamma_j^{(t)}$ ($\gamma_i^{(t)}<\gamma_j^{(t)}$) when $0<\gamma_i^{(0)}<m_i^{(t)}$ ($\gamma_i^{(0)}>m_i^{(t)}$).
If $\alpha_i^{(0)}=\beta_i^{(0)}=0$, then
$\gamma_i^{(t)}=m_i^{(t)}$, and 
 the prior is called an uninformative prior, which is not a proper distribution, although the posterior distribution can be appropriately defined similarly as the informative prior.  
  \\

\noindent \underline{2. Normal Distribution}\\

The conjugate prior for the normal distribution $N(\mu_i,\sigma_i^2)$ with unknown mean and known variance  is a normal distribution $N(\mu_i^{(0)},(\sigma_i^{(0)})^2)$. With (\ref{pd}) and (\ref{prd}), the posterior distribution of $\mu_i$ is $N(\mu_i^{(t)},(\sigma_i^{(t)})^2)$,
where 
\begin{align*}&\mu_i^{(t)}=(\sigma_i^{(t)})^2\left(\frac{\mu_i^{(0)}}{(\sigma_i^{(0)})^2}+\frac{t_i m_i^{(t)}}{\sigma_i^2}\right),\\
 &(\sigma_i^{(t)})^2=\left(\frac{1}{(\sigma_i^{(0)})^2}+\frac{t_i}{\sigma_i^2}\right)^{-1},\end{align*}
and the predictive distribution of $X_{i,t+1}$ is $N(\mu_i^{(t)},\sigma_i^2+(\sigma_i^{(t)})^2)$. If $\sigma_i^{(0)}\to\infty$, $\mu_i^{(t)}=m_i^{(t)}$, and the prior is the uninformative prior in this case. For a normal distribution with unknown variance, there is a normal-gamma conjugate prior (see \cite{degroot2005optimal}). \vspace*{2mm}

 \section{Analysis of Optimal A\&S Policy}\label{asp}
 In Section \ref{os}, we analyze the properties of the optimal selection policy. For discrete sampling and prior distributions, an explicit form for the optimal A\&S policy and its computational complexity are provided in Section \ref{odp}.
 \subsection{Optimal Selection Policy}\label{os}
The optimal selection policy is the last step in the Bellman equation. From (\ref{pd}), we know posterior distributions conditioned on $\mathcal{E}_T$ are independent when the prior distributions for different alternatives are independent, which will be assumed in this section. 
 For PCS, the optimal selection policy is
 \begin{equation*}\label{sp1}
  \begin{aligned}
  \mathcal{S}^{*}(\mathcal{E}_T)&=\arg\max_{i=1,\ldots,k} P(\mu_i\geq \mu_j,~\forall~j\neq i|\mathcal{E}_T)\\
  &=\arg\max_{i=1,\ldots,k}\int_{O_i} \prod_{j\neq i} F_j(x|\mathcal{E}_T) ~f_i(x|\mathcal{E}_T) ~dx,
  \end{aligned}
  \end{equation*}
  where $O_i$ is the feasible set of $\mu_i$, $F_i(\cdot|\mathcal{E}_T)$ is the posterior distribution of $\mu_i$ with density $f_i(\cdot|\mathcal{E}_T)$, $i=1,\ldots,k$, 
 and for EOC, 
 the optimal selection policy is 
  \begin{equation*}\label{sp2}\mathcal{S}^{*}(\mathcal{E}_T)=\arg\max_{i=1,\ldots,k} \mathbb{E}\left[ \mu_i|\mathcal{E}_T\right],\end{equation*}
  and \small
   \begin{align*}
  &V_T(\mathcal{E}_T)=\left.\mathbb{E}\left[ \mu_{i}-\mu_{\langle 1 \rangle}|\mathcal{E}_T\right]\right|_{i=\mathcal{S}^{*}(\mathcal{E}_T)}\\
   =&\left.\mathbb{E}\left[ \mu_i|\mathcal{E}_T\right]\right|_{i=\mathcal{S}^{*}(\mathcal{E}_T)}-\mathbb{E}\left.\left[\sum_{i=1}^{k}\mu_i{\bf1}\{ \mu_i>\mu_j, ~j\neq i \}\right|\mathcal{E}_T\right]\\
  =&\left.\mathbb{E}\left[ \mu_i|\mathcal{E}_T\right]\right|_{i=\mathcal{S}^{*}(\mathcal{E}_T)}-\sum_{i=1}^{k}\mathbb{E}\left.\left[ \mu_i\prod_{j\neq i}\left.\mathbb{E}\left[{\bf1}\{ \mu_i>\mu_j \}\right|\mu_i,\mathcal{E}_T\right]\right|\mathcal{E}_T\right]\\
  =&\left.\mathbb{E}\left[ \mu_i|\mathcal{E}_T\right]\right|_{i=\mathcal{S}^{*}(\mathcal{E}_T)}-\sum_{i=1}^{k}\int_{O_i} x  \prod_{j\neq i} F_j(x|\mathcal{E}_T) ~f_i(x|\mathcal{E}_T) ~dx~.  \end{align*}\normalsize
  
  For EOC, the optimal selection policy for the Bernoulli distribution under conjugacy is 
   $$\mathcal{S}^{*}(\mathcal{E}_T)=\arg\max_{i=1,\ldots,k} \gamma_i^{(T)},$$  
   and the optimal selection policy for the normal distribution under conjugacy is 
   $$\mathcal{S}^{*}(\mathcal{E}_T)=\arg\max_{i=1,\ldots,k} \mu_i^{(T)}~.$$  
    For PCS, the optimal selection policy depends on the entire posterior distributions rather than just the posterior means. 
    For normal distributions with conjugate priors, \cite{peng2014dynamic} showed that except for $\sigma_1^2=\cdots=\sigma_k^2$, selecting the largest posterior mean is not the optimal selection policy, which should also incorporate correlations induced by $\sigma_1^2,\ldots,\sigma_k^2$. 
    
    The following theorem establishes that under some mild conditions, the selection policy selecting the alternative with the largest sample mean is asymptotically consistent with the optimal selection policy for EOC, which is analogous to the result for PCS in  \cite{peng2014dynamic}.

     \begin{theorem} \label{thm2} Suppose for $i=1,\ldots,k$,  $\theta_i=(\mu_i,\xi_i)\in\Omega\times \Xi_i$,   $X_{i,t}\sim Q_i(\cdot;\theta)$,
     i.i.d., $t\in \mathbb{Z}^{+}$, with $Q_i$ mutually independent, $\theta_i\sim F_i(\cdot)$ with $F_i$ mutually independent, and the following conditions are satisfied:
      \begin{itemize}
      \item[(i)]$Q_i(\cdot;\theta)\neq Q_i(\cdot;\theta^{'})$ whenever $\theta\neq\theta^{'}$, $i=1,\ldots,k$;
     \item[(ii)]$P\left( \mu_1=\cdots=\mu_k\right)=0$;
     \item[(iii)] $\mathbb{E}\left[|\mu_i|\right]<\infty$, $i=1,\ldots,k$;
     \item[(iv)]For any $B\subset \Omega$ and finite $T_i$, $P(\mu_i\in B|\mathcal{E}_T)<1$, $i=1,\ldots,k$.
    \end{itemize}
     Then,  we have
     \begin{equation*}\lim_{T\to\infty}\left.\mathbb{E}\left[ V_{E}(\theta;i)|\mathcal{E}_T^{*}\right]\right|_{i=\mathcal{S}^m(\mathcal{E}_T^{*})}=0\quad a.s.,\end{equation*}
      where $\mathcal{S}^m(\mathcal{E}_T^{*})=\arg\max_{i=1,\ldots,k}m_i^{(T)}$ and $\mathcal{E}_T^{*}$ means the information set obtained by following the optimal allocation policy $\mathcal{A}^{*}$, and
     \begin{equation*}
    \lim_{T\to\infty}\left.\mathbb{E}\left[ V_{E}(\theta;i)|\mathcal{E}_T^{*}\right]\right|_{i\neq \mathcal{S}^m(\mathcal{E}_T^{*})}<0\quad a.s.,\end{equation*}
    therefore, $$\lim_{T\to\infty}\left[\mathcal{S}^{*}(\mathcal{E}_T^{*})-\mathcal{S}^{m}(\mathcal{E}_T^{*})\right]=0\quad  a.s.$$ 
     \end{theorem}
      \begin{IEEEproof} Denote $\mathcal{A}^e$ as the equal allocation policy. Following $\mathcal{A}^e$, every alternative will be sampled infinitely often as $n$ goes to infinity. By the law of large numbers, we know
               $$\lim_{T\to\infty}\max_{i=1,\ldots,k} m_i^{(T)}=\max_{i=1,\ldots,k} \mu_i^{*},\quad a.s.,$$
               where $\mu_i^{*}$ means the true parameter. 
                In addition, $\left\{\mathbb{E}[\mu_i|\mathcal{E}_T^e]\right\}$ and $\left\{\mathbb{E}[\max_{i=1,\ldots,k}\mu_i|\mathcal{E}_T^e]\right\}$ are martingales. With condition (iii), we have 
                \begin{align*}&\mathbb{E}\left[\left|\mathbb{E}\left[\mu_i|\mathcal{E}_T^e\right]\right|\right]\leq \mathbb{E}\left[|\mu_i|\right]<\infty,\\
                &\mathbb{E}\left[\left|\mathbb{E}\left[\max_{i=1,\ldots,k}\mu_i|\mathcal{E}_T^e\right]\right|\right]\leq \sum_{i=1}^{k}\mathbb{E}\left[|\mu_i|\right]<\infty,\end{align*}
                where $\mathcal{E}_T^{e}$ means the information set obtained by following  $\mathcal{A}^{e}$.
                By Doob's Martingale Convergence and Consistency Theorems (see \cite{doob1953stochastic} and \cite{van2000asymptotic}), 
                \begin{align*}&\lim_{T\to\infty}\mathbb{E}\left[\mu_i|\mathcal{E}_T^e\right]=\mu_i^{*} ,\\ &\lim_{T\to\infty}\mathbb{E}\left[\max_{i=1,\ldots,k}\mu_i|\mathcal{E}_T^e\right]=\max_{i=1,\ldots,k}\mu_i^{*},\quad a.s.,\end{align*}
                so 
                $$\lim_{T\to\infty}\mathbb{E}\left[ V_{E}(\theta;\mathcal{S}^m(\mathcal{E}_T^e))|\mathcal{E}_T^{e}\right]=0\quad a.s.$$
                By definition, we have 
                  \begin{align*}
                  0=& \lim_{T\to\infty}\mathbb{E}\left[ V_{E}(\theta;\mathcal{S}^m(\mathcal{E}_T^e))|\mathcal{E}_T^{e}\right]\\
                  &\leq \lim_{T\to\infty}\mathbb{E}\left[ V_{E}(\theta;\mathcal{S}^{*}(\mathcal{E}_T^{*}))|\mathcal{E}_T^{*}\right]\leq 0\quad a.s.\end{align*}
                  Then, we prove that following the optimal policy $\mathcal{A}^{*}$, every alternative will be sampled infinitely often  almost surely as $T$ goes to infinity. Otherwise, $\exists$ $k_1,k_2\in\mathbb{Z}^{+}$, s.t. $k_1+k_2=k$ and 
                  \begin{align*}
                  &\left\{ i_1,\ldots,i_{k_1}: T^{*}_{i_{l}}\ed\lim_{T\to\infty} T_{i_{l}}<\infty,~ l=1,\ldots,k_1\right\} \neq \emptyset,\\
                  &\left\{ j_1,\ldots,j_{k_2}: \lim_{n\to\infty} T_{i_{l}}=\infty,~l=1,\ldots,k_2\right\} \neq \emptyset~.\end{align*}
                  We have 
                  \begin{align*}&\mathbb{E}\left[\left|\mathbb{E} \left.\left[ \max_{i=1,\ldots,k} \mu_i\right|\mathcal{E}_T,\mu_{i_1},\ldots,\mu_{i_{k_1}}\right]\right|\right]\\
                  &\leq \mathbb{E}\left[\mathbb{E} \left.\left[ \left|\max_{i=1,\ldots,k} \mu_i\right|\right|\mathcal{E}_T,\mu_{i_1},\ldots,\mu_{i_{k_1}}\right]\right]\leq \sum_{i=1}^{k} \mathbb{E}[|\mu_i|]<\infty~.\end{align*}
                  By the Dominated Convergence Theorem (see \cite{rudin1987real}) and  Doob's Martingale Convergence and Consistency Theorems,\small
                  \begin{align*}
                  &\lim_{T\to\infty}\mathbb{E} \left.\left[ \max_{i=1,\ldots,k} \mu_i\right|\mathcal{E}_T^{*}\right]\\
                  =&\lim_{n\to\infty}\left.\mathbb{E}\left[\mathbb{E} \left.\left[ \max_{i=1,\ldots,k} \mu_i\right|\mathcal{E}_T^{*},\mu_{i_1},\ldots,\mu_{i_{k_1}}\right]\right|\mathcal{E}_T^{*}\right]\\
                  =&\left.\mathbb{E}\left[\lim_{T\to\infty}\mathbb{E} \left.\left[ \max_{i=1,\ldots,k} \mu_i\right|\mathcal{E}_T^{*},\mu_{i_1},\ldots,\mu_{i_{k_1}}\right]\right|\mathcal{E}_T^{*}\right]\\
                  =&\left.\mathbb{E}\left[ \max\left\{ \max_{l=1,..,k_1} \mu_{i_{l}},\max_{l=1,..,k_2}\mu_{j_{l}}^{*}\right \}\right|\bar{X}_{i_1}^{(T)},..,\bar{X}_{i_{k_1}}^{(T)},T^{*}_{i_{1}},..,T^{*}_{i_{k_1}},\zeta_0\right],
                  \end{align*}\normalsize
                  where the last equality holds almost surely. 
                  From the independence condition in the theorem, the conclusion of Theorem \ref{thm1}, and condition (iv),  for $l=1,\ldots,k_1$,
                  \begin{align*}&P\left(\mu_{i_{l}}>C \left|\bar{X}_{i_l}^{(T)}\right.,T_{i_l}^{*},\zeta_0\right)>0,\\ &P\left(\mu_{i_{l}}<C\left|\bar{X}_{i_l}^{(T)}\right.,T_{i_l}^{*},\zeta_0\right)>0,\end{align*}
                  where $C=\max_{l=1,\ldots,k_2}\mu_{j_{l}}^{*}$, so for $l=1,\ldots,k_2$,\small
                  \begin{align*}
                  &\mu_{j_{l}}^{*}-\left.\mathbb{E}\left[ \max\left\{ \max_{l=1,..,k_1} \mu_{i_{l}},\max_{l=1,..,k_2}\mu_{j_{l}}^{*}\right \}\right|\bar{X}_{i_1}^{(T)},..,\bar{X}_{i_{k_1}}^{(T)},T^{*}_{i_{1}},..,T^{*}_{i_{k_1}},\zeta_0\right]\\
                  &\leq \mu_{j_{l}}^{*}-\left.\mathbb{E}\left[ \max\left\{  \mu_{i_{1}},\max_{l=1,\ldots,k_2}\mu_{j_{l}}^{*}\right \}\right|\bar{X}_{i_1}^{(T)},T^{*}_{i_{1}},\zeta_0\right]<0,
                  \end{align*}\normalsize
                  and for $l=1,\ldots,k_1$, 
                  \small
                          \begin{align*}
                          &\mathbb{E}\left[\mu_{i_{l}}\left|\bar{X}_{i_l}^{(T)}\right.,T_{i_t}^{*},\zeta_0\right]\\
                          -&\left.\mathbb{E}\left[ \max\left\{ \max_{l=1,..,k_1} \mu_{i_{l}},\max_{l=1,..,k_2}\mu_{j_{l}}^{*}\right \}\right|\bar{X}_{i_1}^{(T)},..,\bar{X}_{i_{k_1}}^{(T)},T^{*}_{i_{1}},..,T^{*}_{i_{k_1}},\zeta_0\right]\\
                          \leq &\mathbb{E}\left[\mu_{i_{l}}\left|\bar{X}_{i_l}^{(T)}\right.,T_{i_l}^{*},\zeta_0\right]\\
                          &-\left.\mathbb{E}\left[ \max\left\{  \mu_{i_{l}},\max_{l=1,\ldots,k_2}\mu_{j_{l}}^{*}\right \}\right|\bar{X}_{i_l}^{(T)},T^{*}_{i_{l}},\zeta_0\right]<0~.
                          \end{align*}\normalsize
                  Therefore, 
             \begin{align*}
             &\lim_{T\to\infty}\mathbb{E}\left[ V_{E}(\theta;\mathcal{S}^{*}(\mathcal{E}_T^{*}))|\mathcal{E}_T^{*}\right]\\
             &=\max_{i=1,\ldots,k}\mathbb{E}\left[\mu_{i}\left|\mathcal{E}_T^{*}\right.\right]-\mathbb{E}\left.\left[\max_{i=1,\ldots,k}\mu_{i}\right|\mathcal{E}_T^{*}\right]<0\quad a.s.,\end{align*}  
             which is a contradiction.  With every alternative sampled infinitely often under $\mathcal{A}^{*}$, the law of large numbers, and Doob's Martingale Convergence and Consistency Theorems, for $i=\arg\max_{i=1,\ldots,k}\mu_i^{*}$, 
                \begin{align*}\lim_{T\to\infty}\left.\mathbb{E}\left[ V_{E}(\theta;i)|\mathcal{E}_T^{*}\right]\right|_{i=\mathcal{S}^m(\mathcal{E}_T^{*})}=\mu_i^{*}-\max_{i=1,\ldots,k}\mu_i^{*}=0\quad a.s.,\end{align*}  
                and with condition (ii), for $i\neq \arg\max_{i=1,\ldots,k}\mu_i^{*}$, 
                 \begin{align*}\lim_{T\to\infty}\left.\mathbb{E}\left[ V_{E}(\theta;i)|\mathcal{E}_T^{*}\right]\right|_{i\neq \mathcal{S}^m(\mathcal{E}_T^{*})}=\mu_i^{*}-\max_{i=1,\ldots,k}\mu_i^{*}<0\quad a.s.,\end{align*}
                 which completes the proof. 
               \end{IEEEproof}
     \noindent\textbf{Remark.} 
     For independent sampling and prior distributions, the most frequently used conjugate models, including the two models introduced in Section \ref{cj}, satisfy the conditions in Theorem \ref{thm2}  (see \cite{degroot2005optimal}). In the proof, we can see that under mild regularity conditions, every alternative will be sampled infinitely often following the optimal A\&S policy as the simulation budget goes to infinity, which in turn leads to the conclusion of the theorem. 
 \subsection{Optimal A\&S Policy for Discrete Distributions}\label{odp}
 In the case where the sampling distribution and prior distribution are discrete,  in principle the optimal A\&S policy can be calculated. For continuous sampling distribution and prior distribution, discretization can be implemented to reduce to the discrete case as an approximation scheme. 
 Suppose the sampling distribution of $X_{t}$ is supported on $\{ y_{1},\ldots,y_{s} \}$, where $y_j=(y_{1,j},\ldots,y_{k,j})$, $j=1,\ldots,s$, and the prior distribution of $\theta$ is supported on $\{ \eta_{1},\ldots,\eta_{r} \}$. From the results in Theorem \ref{thm1}, the p.m.f. for the posterior distribution of $\theta$ is
 $$f(\eta_{j}|\mathcal{E}_t)=\frac{ \prod_{i=1}^{k}\prod_{\ell=1}^{t_i} q_i(\bar{X}_{i,\ell};\eta_j)~ f(\eta_{j}; \zeta_0)}{\sum_{j^{'}=1}^{r}\prod_{i=1}^{k}\prod_{\ell=1}^{t_i} q_i(\bar{X}_{i,\ell};\eta_{j^{'}})~f(\eta_{j^{'}}; \zeta_0)},$$
 where $f$ is the p.m.f. of joint distribution $F$, 
 and the p.m.f. for the predictive distribution of $X_{i,t+1}$ is
  \begin{align*}&q_i(y_{i,j}|\mathcal{E}_t)\\
  =&\frac{\sum_{j^{'}=1}^{r}q_i(y_{i,j};\eta_{j^{'}})\prod_{i=1}^{k}\prod_{\ell=1}^{t_i} q_i(\bar{X}_{i,\ell};\eta_{j^{'}})~f(\eta_{j^{'}}; \zeta_0)}{\sum_{j^{'}=1}^{r}\prod_{i=1}^{k}\prod_{\ell=1}^{t_i} q_i(\bar{X}_{i,\ell};\eta_{j^{'}})~f(\eta_{j^{'}}; \zeta_0)}~. \end{align*}
 Therefore, 
   $$V_T(\mathcal{E}_T; i)=\mathbb{E}\left[ V(\theta;i)|\mathcal{E}_T \right]=\sum_{j=1}^{r}V(\eta_j;i) ~f(\eta_{j}|\mathcal{E}_T),$$
   and for $0\leq t<T$,
 \begin{align*}
 V_t(\mathcal{E}_t; i)=&\mathbb{E}\left[ V_{t+1}(\mathcal{E}_{t}, X_{i,t+1}) |\mathcal{E}_{t}\right]\\
 =&\sum_{j=1}^{s}V_{t+1}(\mathcal{E}_{t}, y_{i,j}) ~q_i(y_{i,j}|\mathcal{E}_t) , \end{align*}
 so Bellman equations (\ref{be1}) and (\ref{be2}) can be solved recursively.\\     
 
 \noindent\underline{Bernoulli Distribution}\\
 
\begin{figure}[tb]
 \begin{center}
 \begin{picture}(200,180)(20,20)
\put(95,150){$1$}\put(95,45){$0$}\put(95,150){$1$}\put(95,45){$0$}\put(20,110){$0$}\put(190,15){$0$}\put(190,100){$1$}\put(190,180){$2$}
\put(40,55){$q_i(0|0,0)$}\put(40,140){$q_i(1|0,0)$}
\put(105,90){$q_i(1|0,1)$}\put(105,170){$q_i(1|1,1)$}\put(140,125){$q_i(0|1,1)$}\put(140,45){$q_i(0|0,1)$}
 \put(20, 100) {\line(2, 1){80}}
 \put(20, 100) {\line(2, -1){80}}
  \put(100, 60) {\line(2, -1){80}}
   \put(100, 60) {\line(2, 1){80}}
    \put(100, 140) {\line(2, 1){80}}
     \put(100, 140) {\line(2, -1){80}}
 \end{picture}
 \end{center}
 \caption{Evolution of $M_i^{(t)}$, $t=0,1,2$, where $q_i(\cdot| M_i^{(t)},t_i)$ is the predictive probability mass function of $X_{i,t+1}$ conditioned on $(M_i^{(t)},t_i)$}
 % \hspace{1in}
 \label{fig.miv}
 \end{figure}
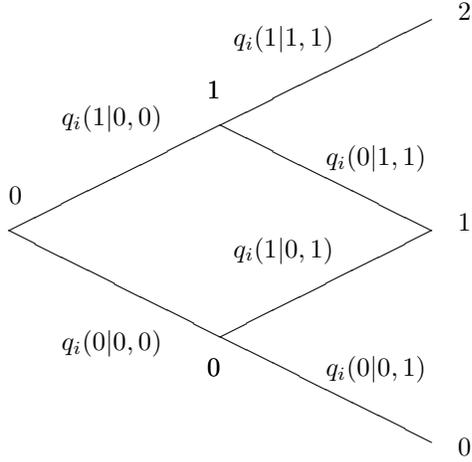

 We analyze the size of the state space for calculating the optimal sampling allocation policy for alternatives following independent Bernoulli distributions, which are the simplest discrete distributions.
 For the Bernoulli distribution, the posterior distribution of alternative $i$ can be determined by the prior information and $(M_i^{(t)},t_i)$, where
 $$M_i^{(t)}\ed\sum_{\ell=1}^{t_i} \bar{X}_{i,\ell}~.$$
 Notice that the number of possible outcomes of $M_i^{(t)}$ is $t_i+1$, which grows linearly with respect to the number of allocated replications. Figure \ref{fig.miv} provides an illustration for the evolution of $M_i^{(t)}$. 
 
 With given prior information $\zeta_0$, $(M_1^{(t)},\ldots,M_k^{(t)},t_1,\ldots,t_k)$ determines the state space of information set $\mathcal{E}_t$. The size of the state space is $$L_{t,k}=\sum_{\left\{(t_1,\ldots,t_k):~t_1+\cdots+t_k=t\right\}}(t_1+1)\times \cdots\times (t_k+1)~.$$  
 To shed some light on how large the state space is, we provide upper and lower bounds for the summation. It is easy to see that 
 $$L_{t,k}\geq \left(\lceil\frac{t}{k}\rceil+1\right)^k,$$
 and 
 $$(t_1+1)\times \cdots\times (t_k+1)\leq (t+1)^k~.$$
The elements in the set $\left\{(t_1,\ldots,t_k):~t_1+\cdots+t_k=t\right\}$ are in one-to-one correspondence to possible choices for picking $k-1$ balls from $t+k-1$ balls.  
 From simple combinatorics, the size of the set $\left\{(t_1,\ldots,t_k):~t_1+\cdots+t_k=t\right\}$ is $$C_{t+k-1}^{k-1}=\frac{(t+k-1)!}{t!(k-1)!},$$
thus
 $$L_{t,k}\leq \frac{(t+k-1)^{2k}}{(k-1)!}~.$$
For a fixed $k$, $L_{t,k}$ grows at a polynomial rate with respect to $t$. However, if $t$ and $k$ grow together and $t>k$, the lower bound of $L_{t,k}$ grows at an exponential rate with respect to $k$.
 From $T$ to $1$, we can use backward induction to determine the optimal allocation policy $A^{*}_{t}(\cdot)$ for every possible state of $\mathcal{E}_{t-1}$. \\
 
  \noindent\underline{General Discrete Distribution}\\
  
 The size of the state space for general independent discrete sampling distributions can be analyzed similarly as in  the case of independent Bernoulli distributions.  Since the product operation is commutative in the likelihood, a possible outcome  $(\bar{X}_{i,1},\ldots,\bar{X}_{i,t_i})$ leading to a distinctive posterior distribution is uniquely determined by the number of elements picked from $(y_{i,1},\ldots,y_{i,s_i})$. The size of the possible outcomes is equivalent to the size of the set $\{ (c_1,\ldots,c_{s_i})\in\mathbb{N}^{s_i}: c_1+\cdots+ c_{s_i}=t_i \}$, which is 
 $$C_{s_i+t_i-1}^{s_i-1}=\frac{(s_i+t_i-1)!}{t_i!(s_i-1)!}~.$$
 The size of the state space for the information set  $\mathcal{E}_t$ is 
 $$L_{t,k}=\sum_{\left\{(t_1,\ldots,t_k):~t_1+\cdots+t_k=t\right\}}\prod_{i=1}^{k}C_{s_i+t_i-1}^{s_i-1}~.$$
 Denote $\overline{s}\ed\max_{i=1}^k s_i$ and $\underline{s}\ed\min_{i=1}^k s_i$. Then, we have 
 $$ \left(1+\lceil\frac{t}{k}\rceil/(\overline{s}-1)\right)^{k(\underline{s}-1)}\leq L_{t,k}\leq \frac{(\overline{s}+t+k-1)^{k\overline{s}}}{(\overline{s}-1)!(k-1)!}~.$$
 Similarly, for fixed $k$ and $\overline{s}$, $L_{t,k}$ grows at a polynomial rate with respect to $t$. However, if $t$, $k$ and $\underline{s}$  grow together and $t>k(\overline{s}-1)$, the lower bound of $L_{t,k}$ grows at an exponential rate with respect to $k$ and $\underline{s}$.
 \section{A\&S Policy for Normal Distributions}\label{asn}
 In this section, we consider the sampling allocation for alternatives following independent normal distributions, which is most frequently assumed in  R\&S, and we derive an efficient scheme to approximate the optimal A\&S policy in an ADP paradigm. More specifically, we do not apply backward induction, but use forward programming by optimizing a value function approximation (VFA) one step ahead. Similar to many ADP approaches  (see \cite{powell2007approximate}), the proposed procedure in this work is not guaranteed to achieve the optimal A\&S policy, but can alleviate the curse of dimensionality.    

 We focus on PCS as the final reward of the selection decision and assume a conjugate prior for the unknown means in Section \ref{cj}. In this section, we suppose any step $t$ could be the last step.
  The selection policy is to choose the alternative with the largest posterior mean, which is asymptotically optimal based on analysis in Section \ref{os}, i.e.,
 $$\widehat{\mathcal{S}}(\mathcal{E}_t)=\langle 1 \rangle_t,$$
 where $\mu_{\langle 1 \rangle_t}^{(t)}>\cdots>\mu_{\langle k\rangle_t}^{(t)}$.
 As in ADP, we approximate the value function using some features, specifically, the following VFA is used for selecting the $\langle 1 \rangle_t$-th alternative:
 \begin{align}\label{nn}\bar{V}(\mathcal{E}_t;w)=K\left(\sum_{i=1}^{\tau} w_i g_i(\mathcal{E}_t)\right),\end{align}
 where $g_j(\cdot)$, $j=1,\ldots,k$, are features of the value function, $w=(w_1,\ldots,w_{\tau})$ are the weights of the features, and $K(\cdot)$ is referred to as the activation function. 
 
  \subsection{Approximately Optimal Allocation Policy}
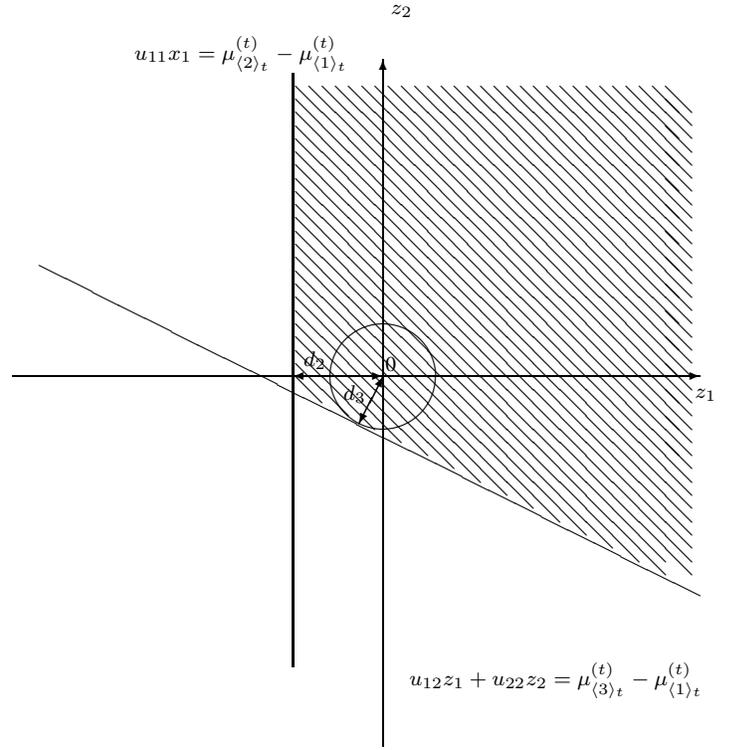
\begin{figure}[tb]
\footnotesize
\begin{center}
\begin{picture}(280,280)(20,0)
\put (20, 140) {\vector(1, 0){260}} \put (160, 0) {\vector(0,
1){260}} \put (278, 132) {$z_1$} \put (163, 277) {$z_2$} \put
(161, 142) {0} \put (126, 30) {\line(0, 1){225}} \put (66, 260)
{$u_{11}x_1=\mu_{\langle 2\rangle_t}^{(t)}-\mu_{\langle 1 \rangle_t}^{(t)}$} 
\put (30, 182) {\line(2, -1){250}} \put
(170, 23) {$u_{12}z_1+u_{22}z_2=\mu_{\langle 3\rangle_t}^{(t)}-\mu_{\langle 1 \rangle_t}^{(t)}$} \put (143, 140)
{\vector(-1, 0){17}} \put (143, 140) {\vector(1, 0){17}} \put
(130, 144) {$d_2$} \put (160, 140) {\vector(-1, -2){9}} \put (151,
122) {\vector(1, 2){9}} \put (145, 131) {$d_3$} \put (160, 140)
{\circle{40}}
%
% shading
\put (127,140) {\line(1,-1){10}} \put (127,145) {\line(1,-1){20}}
\put (127,150) {\line(1,-1){30}} \put (127,155) {\line(1,-1){40}}
\put (127,160) {\line(1,-1){50}} \put (127,165) {\line(1,-1){60}}
\put (127,170) {\line(1,-1){70}} \put (127,175) {\line(1,-1){80}}
\put (127,180) {\line(1,-1){90}} \put (127,185) {\line(1,-1){100}}
\put (127,190) {\line(1,-1){110}} \put (127,195)
{\line(1,-1){120}} \put (127,200) {\line(1,-1){130}} \put
(127,205) {\line(1,-1){140}} \put (127,210) {\line(1,-1){140}}
\put (127,215) {\line(1,-1){150}} \put (127,220)
{\line(1,-1){150}} \put (127,225) {\line(1,-1){150}} \put
(127,230) {\line(1,-1){150}} \put (127,235) {\line(1,-1){150}}
\put (127,240) {\line(1,-1){150}} \put (127,245)
{\line(1,-1){150}} \put (127,250) {\line(1,-1){150}} \put
(132,250) {\line(1,-1){145}} \put (137,250) {\line(1,-1){140}}
\put (142,250) {\line(1,-1){135}} \put (147,250)
{\line(1,-1){130}} \put (152,250) {\line(1,-1){125}} \put
(157,250) {\line(1,-1){120}} \put (162,250) {\line(1,-1){115}}
\put (167,250) {\line(1,-1){110}} \put (172,250)
{\line(1,-1){105}} \put (177,250) {\line(1,-1){100}} \put
(182,250) {\line(1,-1){95}} \put (187,250) {\line(1,-1){90}}
\put (192,250) {\line(1,-1){85}} \put (197,250)
{\line(1,-1){80}} \put (202,250) {\line(1,-1){75}} \put (207,250)
{\line(1,-1){70}} \put (212,250) {\line(1,-1){65}} \put (217,250)
{\line(1,-1){60}} \put (222,250) {\line(1,-1){55}} \put (227,250)
{\line(1,-1){50}} \put (232,250) {\line(1,-1){45}} \put (237,250)
{\line(1,-1){40}} \put (242,250) {\line(1,-1){35}} \put (247,250)
{\line(1,-1){30}} \put (252,250) {\line(1,-1){25}} \put (257,250)
{\line(1,-1){20}} \put (262,250) {\line(1,-1){15}} \put (267,250)
{\line(1,-1){10}} \put (272,250) {\line(1,-1){5}}
 %\put (277,250){\line(1,-1){5}} 
%\put (282,250) {\line(1,-1){15}}
 %\put (287,250){\line(1,-1){10}}
%
\end{picture}
\end{center}
\caption{Area of integration for approximation is the circle,
where dominant values of integrand $\exp(-(z_1^2+z_2^2)/2)$
are captured.}
% \hspace{1in}
\label{fig.approx}
\end{figure}

   We first provide a VFA using one feature in the value function.
The  value function  for selecting the $\langle 1 \rangle_t$-th alternative is 
 \begin{align}\label{vf}
 \mathbb{E}[V_P(\theta;\langle 1 \rangle_t)| \mathcal{E}_t]=P\left( \mu_{\langle 1 \rangle_t}>\mu_{\langle i \rangle_t},~i\neq 1|\mathcal{E}_t\right),
 \end{align}
 which is the posterior integrated PCS (IPCS). 
 Conditioned on $\mathcal{E}_t$, $\mu_i$ follows a normal distribution with mean $\mu_i^{(t)}$ and variance $(\sigma_i^{(t)})^2$, $i=1,\ldots,k$. Therefore, the joint distribution of vector $(\mu_{\langle 1 \rangle_t}-\mu_{\langle 2\rangle_t},\ldots,\mu_{\langle 1 \rangle_t}-\mu_{\langle k\rangle_t})$ follows a joint normal distribution with mean $(\mu_{\langle 1 \rangle_t}^{(t)}-\mu_{\langle 2\rangle_t}^{(t)},\ldots,\mu_{\langle 1 \rangle_t}^{(t)}-\mu_{\langle k\rangle_t}^{(t)})$ and covariance matrix $\Gamma^{'}~\Xi ~\Gamma$, where $'$ indicates the transpose operation, 
 $ \Lambda=diag( (\sigma_{\langle 1 \rangle_t}^{(t)})^2,\ldots,(\sigma_{\langle k\rangle_t}^{(t)})^2),
 $
and 
$$
\Gamma = \left( \begin{array}{rrrrr}
1 & 1 & ~~1 & ~~\cdots & 1 \\
-1 & 0 & 0 & \cdots & 0 \\
0 & -1 & 0 & \cdots & 0 \\
\vdots & \vdots & \vdots & \cdots & \vdots \\
0 & 0 & 0 & \cdots & -1
\end{array} \right)_{k \times (k-1)} .
$$
By Cholesky decomposition $\Gamma^{'}~\Xi ~\Gamma=U^{'} U$, where $U=[u_{ij}]_{(k-1)\times (k-1)}$ is an upper triangular matrix
(i.e., $u_{ij}=0$ if $i>j$),
\begin{align*}
& P(\mu_{\langle 1 \rangle_t}>\mu_{\langle i \rangle_t},~i\neq 1|\mathcal{E}_t) \\=&P\left(\sum_{j=1}^{i-1} u_{j,i-1} Z_{j} > \mu_{\langle i \rangle_t}^{(t)} - \mu_{\langle 1 \rangle_t}^{(t)}, ~i\neq 1\right)=\frac{1}{(\sqrt{2\pi})^{k-1}}  \\
 &\times\iint_{\sum_{j=1}^{k-1} u_{j,i} z_{j} \geq \mu_{\langle i \rangle_t}^{(t)}- \mu_{\langle 1 \rangle_t}^{(t)}}\prod_{\ell=1}^{k-1}\exp\left(-\frac{z_{j}^2}{2}\right)d z_{k-1} \cdots dz_1,
\end{align*}
where $Z_i$, $i=1,\ldots,k-1$, are independent standard normal random variables. The value function  (\ref{vf}) is an integration of the density of a $(k-1)$-dimensional standard normal distribution over an area covered by hyperplanes: $$\sum_{j=1}^{i}
u_{j,i}z_j = \mu_{\langle 1 \rangle_t}^{(t)}-\mu_{\langle 1 \rangle_t}^{(t)},\quad i\neq 1.$$
Since the density of the normal distribution decreases at an exponential rate with respect to the distance from the origin, the size of the hypersphere centered at the origin can capture the magnitude of the integration over the whole area covered by the hyperplanes. 
A feature of the  value function  (\ref{vf}) is the size of the hypersphere, i.e.,
 $$\pi d^2(\mathcal{E}_t)/2,\quad s.t. ~d(\mathcal{E}_t)=\min \left(d_{2}(\mathcal{E}_t), d_{3}(\mathcal{E}_t), \ldots, d_{k}(\mathcal{E}_t)\right),$$ where $d_{i}$
is the distance from the origin to hyperplane, which is visualized for $k=3$ in Figure \ref{fig.approx}. 
Simple algebra yields 
$$d_i(\mathcal{E}_t) =\frac{\mu_{\langle 1 \rangle_t}^{(t)}-\mu_{\langle i \rangle_t}^{(t)}}{\sqrt{\left(\sigma_{\langle 1 \rangle_t}^{(t)}\right)^2+\left(\sigma_{\langle i \rangle_t}^{(t)}\right)^2}}~.$$
We use a VFA for the value function (\ref{vf}) given by $\widetilde{V}_t(\mathcal{E}_t)=d^2(\mathcal{E}_t)$. 
If the $(t+1)$-th replication is the last one, a VFA looking one step ahead at step $t$ by allocating the $i$th alternative can be given as follows:
\begin{equation*}\widetilde{V}_t(\mathcal{E}_t;i)=\left.\mathbb{E}\left[ \widetilde{V}_{t+1}(\mathcal{E}_{t}, X_{i,t+1})\right|\mathcal{E}_{t}\right]~.\end{equation*}
Since the above expectation is difficult to calculate, we use the following certainty equivalence (\cite{bertsekas2005dynamic}) as an approximation: for $ j\neq 1$,
\begin{equation}\label{ax}
\begin{aligned}
 &\widehat{V}_{t}(\mathcal{E}_{t};1)\ed\widetilde{V}_{t+1}\left(\mathcal{E}_{t}, \left.\mathbb{E}\left[X_{1,t+1}\right|\mathcal{E}_{t}\right]\right)\\
 &\qquad\quad~=\min_{i\neq 1}\frac{\left(\mu_{\langle 1 \rangle_t}^{(t)}-\mu_{\langle i \rangle_t}^{(t)}\right)^2}{\left(\sigma_{\langle 1 \rangle_t}^{(t+1)}\right)^2+\left(\sigma_{\langle i \rangle_t}^{(t)}\right)^2},\\
 &\widehat{V}_{t}(\mathcal{E}_{t};j)\ed\widetilde{V}_{t+1}\left(\mathcal{E}_{t}, \left.\mathbb{E}\left[X_{j,t+1}\right|\mathcal{E}_{t}\right]\right)\\
 &=\min\left\{\frac{\left(\mu_{\langle 1 \rangle_t}^{(t)}-\mu_{\langle j\rangle_t}^{(t)}\right)^2}{\left(\sigma_{\langle 1 \rangle_t}^{(t)}\right)^2+\left(\sigma_{\langle j\rangle_t}^{(t+1)}\right)^2},~\min_{i\neq 1,j}\frac{\left(\mu_{\langle 1 \rangle_t}^{(t)}-\mu_{\langle i \rangle_t}^{(t)}\right)^2}{\left(\sigma_{\langle 1 \rangle_t}^{(t)}\right)^2+\left(\sigma_{\langle i \rangle_t}^{(t)}\right)^2}\right\}.
\end{aligned}
\end{equation}
An AOAP that optimize the VFA one step ahead is given by 
\begin{equation}\label{map}\widehat{A}_{t+1}(\mathcal{E}_{t})=\arg\max_{i=1,\ldots,k}\widehat{V}_{t}(\mathcal{E}_{t};i)~.\end{equation}

In contrast with the approximation scheme in \cite{fu2007simulation} and \cite{peng2012efficient}, which used $d$ as the objective function of a surrogate parametric optimization,  we use $d^2$ as the VFA to obtain a sequential policy. This choice of $d^2$ instead of $d$ will lead to a desirable asymptotic property of the proposed AOAP  (\ref{map}). 
In contrast with knowledge gradient (KG), which uses a surrogate $\mu_{\langle 1 \rangle_{t+1}}^{(t+1)}$ in the derivation of the one-step-ahead optimality (\cite{gupta1996bayesian}, \cite{frazier2008knowledge}),  the VFA in AOAP (\ref{map}) uses an easily interpretable feature (size of hypersphere) that better describes the true value function in our SCP.  AOAP (\ref{map}) possesses the following asymptotic properties. 

\begin{theorem} \label{thm3} AOAP (\ref{map}) is consistent, i.e.,
$$\lim_{t\to\infty}\widehat{\mathcal{S}}(\mathcal{E}_t)=\langle 1 \rangle\quad a.s.$$
In addition, the sampling ratio of each alternative asymptotically achieves the optimal decreasing rate of the large deviations of the probability of false selection in \cite{glynn2004large}, i.e., 
$$\lim_{t\to\infty} r_i^{(t)}= r_i^{*},\quad a.s.,\quad i=1,\ldots,k,$$
where $r_i^{(t)}\ed t_i/t$, $\sum_{i=1}^{k} r_i^{*}=1$, $r_i^{*}\geq 0$, $i=1,\ldots,k$, and for $i,j\neq 1$, 
\begin{align}
&\frac{(\mu_{\langle i \rangle}-\mu_{\langle 1 \rangle})^2}{\left(\sigma_{\langle i \rangle}\right)^2/r_{\langle i \rangle}^{*}+\left(\sigma_{\langle 1 \rangle}\right)^2/r_{\langle 1 \rangle}^{*}}=\frac{(\mu_{\langle j\rangle}-\mu_{\langle 1 \rangle})^2}{\left(\sigma_{\langle j\rangle}\right)^2/r_{\langle j\rangle}^{*}+\left(\sigma_{\langle 1 \rangle}\right)^2/r_{\langle 1 \rangle}^{*}},\label{a1}\\ \label{a2}
&r_{\langle 1 \rangle}^{*}=\left(\sigma_{\langle i \rangle}\right)^2\sqrt{\sum_{i\neq 1}\left(r_{\langle i \rangle}^{*}\right)^2/\left(\sigma_{\langle i \rangle}\right)^2}~.
\end{align}
\end{theorem}
\begin{IEEEproof} We only need to prove that every alternative will be sampled infinitely often almost surely, following  AOAP (\ref{map}), and the consistency will follow by the law of large numbers. 
   Suppose alternative $i$ is only sampled finitely often and alternative $j$ is sampled infinitely often. Then, 
   $$\lim_{t\to\infty} \widehat{V}_t(\mathcal{E}_t;i)-\widetilde{V}_t(\mathcal{E}_t)>0,~ \lim_{t\to\infty} \widehat{V}_t(\mathcal{E}_t;j)-\widetilde{V}_t(\mathcal{E}_t)=0,~ a.s.,$$ 
   which contradicts with the sampling rule that the alternative with the largest VFA is sampled in AOAP (\ref{map}). Therefore, AOAP (\ref{map}) must be consistent. 
   
   By the law of large numbers, $\lim_{t\to\infty}\mu_{i}^{(t)}\to\mu_i$, $i=1,\ldots,k$.
    Because the asymptotic sampling ratios will be determined by the increasing order of $\widehat{V}_t(\mathcal{E}_t;i)$ with respect to $t$, we can replace $\mu_i^{(t)}$ and $(\sigma_i^{(t)})^2$ with $\mu_i$ and $\sigma_i^{2}/t_i$ in $\widehat{V}_t(\mathcal{E}_t;i)$, $i=1,\ldots,k$, for simplicity of analysis. If $(r_1^{(t)},\ldots,r_k^{(t)})$ does not converges to $(r_1^{*},\ldots,r_k^{*})$, there exists a subsequence of the former converging to $(\widetilde{r}_1,\ldots,\widetilde{r}_k)\neq (r_1^{*},\ldots,r_k^{*})$ such that  $\sum_{i=1}^{k} \widetilde{r}_i=1$, $\widetilde{r}_i\geq 0$, $i=1,\ldots,k$,  by the Bolzano-Weierstrass theorem (\cite{rudin1964principles}). Without loss of generality, we can assume $(r_1^{(t)},\ldots,r_k^{(t)})$ converges to $(\widetilde{r}_1,\ldots,\widetilde{r}_k)$. 
    We claim $\widetilde{r}_i>0$, $i=1,\ldots,k$; otherwise, there exists $\widetilde{r}_{\langle j\rangle}=0$ and  $\widetilde{r}_{[j^{'}]}>0$. Notice
    \begin{align*}
    &\lim_{t\to\infty}\left[\frac{\left(\mu_{\langle 1 \rangle}-\mu_{\langle i \rangle}\right)^2}{\sigma_{\langle 1 \rangle}^2/t_{\langle 1 \rangle}+\sigma_{\langle i \rangle}^2/(t_{\langle i \rangle}+1)}-\frac{\left(\mu_{\langle 1 \rangle}-\mu_{\langle i \rangle}\right)^2}{\sigma_{\langle 1 \rangle}^2/t_{\langle 1 \rangle}+\sigma_{\langle i \rangle}^2/t_{\langle i \rangle}}\right]\\
    =&\lim_{t\to\infty}t\left[\frac{\left(\mu_{\langle 1 \rangle}-\mu_{\langle i \rangle}\right)^2}{\sigma_{\langle 1 \rangle}^2/r^{(t)}_{\langle 1 \rangle}+\sigma_{\langle i \rangle}^2/(r^{(t)}_{\langle i \rangle}+1/t)}-\frac{\left(\mu_{\langle 1 \rangle}-\mu_{\langle i \rangle}\right)^2}{\sigma_{\langle 1 \rangle}^2/r^{(t)}_{\langle 1 \rangle}+\sigma_{\langle i \rangle}^2/r^{(t)}_{\langle i \rangle}}\right]\\
    =& \lim_{t\to\infty}\left.\frac{\partial G_i(r^{(t)}_{\langle 1 \rangle},x)}{\partial x}\right|_{x=r_{\langle i \rangle}^{(t)}}\\
    &=\lim_{t\to\infty} \left(\frac{\sigma_{\langle i \rangle}}{r_{\langle i \rangle}^{(t)}}\right)^2 \frac{(\mu_{\langle i \rangle}-\mu_{\langle 1 \rangle})^2}{\left(\sigma_{\langle i \rangle}^2/r_{\langle i \rangle}^{(t)}+\sigma_{\langle 1 \rangle}^2/r_{\langle 1 \rangle}^{(t)}\right)^2},
    \end{align*}
    and 
     \begin{align*}
     &\lim_{t\to\infty}\left[\frac{\left(\mu_{\langle 1 \rangle}-\mu_{\langle i \rangle}\right)^2}{\sigma_{\langle 1 \rangle}^2/(t_{\langle 1 \rangle}+1)+\sigma_{\langle i \rangle}^2/t_{\langle i \rangle}}-\frac{\left(\mu_{\langle 1 \rangle}-\mu_{\langle i \rangle}\right)^2}{\sigma_{\langle 1 \rangle}^2/t_{\langle 1 \rangle}+\sigma_{\langle i \rangle}^2/t_{\langle i \rangle}}\right]\\
     =&\lim_{t\to\infty}t\left[\frac{\left(\mu_{\langle 1 \rangle}-\mu_{\langle i \rangle}\right)^2}{\sigma_{\langle 1 \rangle}^2/\left(r^{(t)}_{\langle 1 \rangle}+1/t\right)+\sigma_{\langle i \rangle}^2/r^{(t)}_{\langle i \rangle}}-\frac{\left(\mu_{\langle 1 \rangle}-\mu_{\langle i \rangle}\right)^2}{\sigma_{\langle 1 \rangle}^2/r^{(t)}_{\langle 1 \rangle}+\sigma_{\langle i \rangle}^2/r^{(t)}_{\langle i \rangle}}\right]\\
     =& \lim_{t\to\infty}\left.\frac{\partial G_i(x,r^{(t)}_{i})}{\partial x}\right|_{x=r_{\langle 1 \rangle}^{(t)}}\\  
     &=\lim_{t\to\infty} \left(\frac{\sigma_{\langle 1 \rangle}}{r_{\langle 1 \rangle}^{(t)}}\right)^2 \frac{(\mu_{\langle i \rangle}-\mu_{\langle 1 \rangle})^2}{\left(\sigma_{\langle i \rangle}^2/r_{\langle i \rangle}^{(t)}+\sigma_{\langle 1 \rangle}^2/r_{\langle 1 \rangle}^{(t)}\right)^2},
     \end{align*}
    where 
    $$G_i(r_{\langle 1 \rangle},r_{\langle i \rangle})\ed\frac{(\mu_{\langle i \rangle}-\mu_{\langle 1 \rangle})^2}{\sigma_{\langle i \rangle}^2/r_{\langle i \rangle}+\sigma_{\langle 1 \rangle}^2/r_{\langle 1 \rangle}},\quad i\neq 1.$$
    We have 
    \begin{align*}&\lim_{t\to\infty} \left(\frac{\sigma_{\langle j\rangle}}{r_{\langle j\rangle}^{(t)}}\right)^2 \frac{(\mu_{\langle i \rangle}-\mu_{\langle 1 \rangle})^2}{\left(\sigma_{\langle i \rangle}^2/r_{\langle i \rangle}^{(t)}+\sigma_{\langle 1 \rangle}^2/r_{\langle 1 \rangle}^{(t)}\right)^2}=\infty,\\
    &\lim_{t\to\infty} \left(\frac{\sigma_{\langle j^{'}\rangle}}{r_{\langle j^{'}\rangle}^{(t)}}\right)^2 \frac{(\mu_{\langle i^{'}\rangle}-\mu_{\langle 1 \rangle})^2}{\left(\sigma_{\langle i^{'}\rangle}^2/r_{\langle i^{'}\rangle}^{(t)}+\sigma_{\langle 1 \rangle}^2/r_{\langle 1 \rangle}^{(t)}\right)^2}<\infty, \end{align*}
    where $j=i$ or $j=1$, and $j^{'}=i^{'}$ or $j^{'}=1$. This  contradicts  the sampling rule of AOAP (\ref{map}), so $\widetilde{r}_i>0$, $i=1,\ldots,k$.  
    If $(\widetilde{r}_1,\ldots,\widetilde{r}_k)$ does not satisfy (\ref{a1}), there exist $i\neq j$, $i,j\neq 1$ such that 
    $$G_i(\widetilde{r}_{\langle 1 \rangle},\widetilde{r}_{\langle i \rangle})>G_j(\widetilde{r}_{\langle 1 \rangle},\widetilde{r}_{\langle j\rangle})~.$$
    If the inequality above holds, there exists $T_0>0$ such that\\
     $\forall$ $t>T_0$, 
     $$G_i(r^{(t)}_{\langle 1 \rangle},r^{(t)}_{\langle i \rangle})>G_j(r^{(t)}_{\langle 1 \rangle},r^{(t)}_{\langle j\rangle}),$$
     due to the continuity of $G_i$ and $G_j$ on $(0,1)\times(0,1)$. By the  sampling rule of AOAP (\ref{map}), the $\langle j\rangle$th alternative will be sampled and the $\langle i \rangle$th alternative will stop receiving replications before the inequality above reverses. This contradicts $(r_1^{(t)},\ldots,r_k^{(t)})$ converging to $(\widetilde{r}_1,\ldots,\widetilde{r}_k)$, so (\ref{a1}) must hold. By the implicit function theorem (\cite{rudin1964principles}), (\ref{a1}) and $\sum_{i=1}^{k} \widetilde{r}_i=1$ determine implicit functions  $\widetilde{r}_{\langle i \rangle}(x)|_{x=\widetilde{r}_{\langle 1 \rangle}}$, $i=2,\ldots,k$, because 
     \begin{align*}
     \det(\Sigma)=\prod_{i=2}^{k}\zeta_{i,i}
     \left\{\sum_{i=2}^{k}\zeta_{i,i}^{-1}\right\}>0, \end{align*}
      where    
      $$\zeta_{i,i}\ed\left.\frac{\partial G_i(\widetilde{r}_{\langle 1 \rangle},x)}{\partial x}\right|_{x=\widetilde{r}_{\langle i \rangle}},\qquad i=2,\ldots,k,$$
          $$
           \Sigma \ed\left( \begin{array}{cccccc}
           \zeta_{2,2}& -\zeta_{3,3} & \cdots & 0&0\\
           0& \zeta_{3,3}& \cdots & 0&0 \\
           \vdots  & \vdots & \cdots & \vdots &\vdots\\
            0 & 0 & \cdots & \zeta_{k-1,k-1}&-\zeta_{k,k}\\
           1 & 1 & \cdots & 1& 1
           \end{array} \right),
           $$
     %\begin{align*}
     %\Sigma &\ed\left( \begin{array}{cccc}
     %\left.\frac{\partial G_2(\widetilde{r}_{\langle 1 \rangle},x)}{\partial x}\right|_{x=\widetilde{r}_{\langle 2\rangle}}& -\left.\frac{\partial G_3(\widetilde{r}_{\langle 1 \rangle},x)}{\partial x}\right|_{x=\widetilde{r}_{\langle 3\rangle}} & \cdots \\
     %0& \left.\frac{\partial G_3(\widetilde{r}_{\langle 1 \rangle},x)}{\partial x}\right|_{x=\widetilde{r}_{\langle 3\rangle}} & \cdots \\
     %\vdots  & \vdots & \cdots \\
     % 0 & 0 & \cdots \\
     %1 & 1 & \cdots 
     %\end{array} \right.\\
     %&\qquad\left. \begin{array}{ccc}
      %   0&0\\
       %    0&0 \\
        %  \vdots &\vdots\\
        %   \left.\frac{\partial G_{k-1}(\widetilde{r}_{\langle 1 \rangle},x)}{\partial x}\right|_{x=\widetilde{r}_{\langle k-1\rangle}}&-\left.\frac{\partial G_k(\widetilde{r}_{\langle 1 \rangle},x)}{\partial x}\right|_{x=\widetilde{r}_{\langle k\rangle}}\\
        %  1& 1
         % \end{array} \right),
     %\end{align*}
     and $R=-\Sigma^{-1} \Upsilon$, where 
     \begin{align*}
     &R\ed\left.\left(\frac{\partial \widetilde{r}_{\langle 2\rangle}(x)}{\partial x},\ldots,\frac{\partial \widetilde{r}_{\langle k\rangle}(x)}{\partial x}\right)^{'}\right|_{x= \widetilde{r}_{\langle 1 \rangle}},\\
     &\Upsilon\ed\left(\frac{\partial G_2(x,\widetilde{r}_{\langle 2\rangle})}{\partial x}-\frac{\partial G_3(x,\widetilde{r}_{\langle 3\rangle})}{\partial x},\right.\\
     &\left.\left.\ldots,\frac{\partial G_{k-1}(x,\widetilde{r}_{\langle k-1\rangle})}{\partial x}-\frac{\partial G_k(x,\widetilde{r}_{\langle k\rangle})}{\partial x},1\right)^{'}\right|_{x=\widetilde{r}_{\langle 1 \rangle}}.
     \end{align*}
     In addition,  
     $$\left.\frac{\partial G_i(x,\widetilde{r}_{\langle i \rangle})}{\partial x}\right|_{x=\widetilde{r}_{\langle 1 \rangle}}+\zeta_{i,i}\left.\frac{\partial \widetilde{r}_{\langle i \rangle}(x)}{\partial x}\right|_{x=\widetilde{r}_{\langle 1 \rangle}}=0,\quad i\neq 1;$$
     otherwise, there exists $j\neq 1$ such that the equality above does not hold, say 
      $$\left.\frac{\partial G_j(x,\widetilde{r}_{\langle j\rangle})}{\partial x}\right|_{x=\widetilde{r}_{\langle 1 \rangle}}+\zeta_{j,j}\left.\frac{\partial \widetilde{r}_{\langle j\rangle}(x)}{\partial x}\right|_{x=\widetilde{r}_{\langle 1 \rangle}}>0~.$$
      Following the sampling rule of AOAP (\ref{map}), the $\langle 1 \rangle$th alternative will be sampled and the $\langle j\rangle$th alternative will stop receiving replications before the inequality above is no longer satisfied, which contradicts $(r_1^{(t)},\ldots,r_k^{(t)})$ converging to $(\widetilde{r}_1,\ldots,\widetilde{r}_k)$. Then, $H R=-G$, where 
      $$G\ed\left.\left(\frac{\partial G_2(x,\widetilde{r}_{\langle 2\rangle})}{\partial x},\ldots,\frac{\partial G_k(x,\widetilde{r}_{\langle k\rangle})}{\partial x}\right)^{'}\right|_{x=\widetilde{r}_{\langle 1 \rangle}}$$
      and
        $$
       H\ed\left( \begin{array}{cccc}
       \zeta_{2,2} & 0&\cdots & 0\\
       0 & \zeta_{3,3}&\cdots & 0\\
        \vdots& \vdots& \ddots & \vdots \\
         0 & \cdots&0&\zeta_{k,k}        \end{array} \right)~.
        $$
        Summarizing the above, we have 
        $$\Upsilon=\Sigma~ H^{-1} G,$$
        which leads to 
        $$\sum_{i=2}^{k}\frac{\left.\partial G_i(x,\widetilde{r}_{\langle i \rangle})/\partial x\right|_{x=\widetilde{r}_{\langle 1 \rangle}}}{\left.\partial G_i(\widetilde{r}_{\langle 1 \rangle},x)/\partial x\right|_{x=\widetilde{r}_{\langle i \rangle}}}=1~\Leftrightarrow~ (\ref{a2})~.$$
   Since there is only one solution to (\ref{a1}) and (\ref{a2}) (see \cite{glynn2004large}), we have 
   $(\widetilde{r}_1,\ldots,\widetilde{r}_k)=(r_1^{*},\ldots,r_k^{*})$, which contradicts the premise that $(r_1^{(t)},\ldots,r_k^{(t)})$ does not converges to $(r_1^{*},\ldots,r_k^{*})$. Therefore, $(r_1^{(t)},\ldots,r_k^{(t)})$  converges to $(r_1^{*},\ldots,r_k^{*})$, which proves the theorem. 
 \end{IEEEproof}
\noindent\textbf{Remark.} From \cite{glynn2004large}, if $r_{\langle 1 \rangle}^{*}\gg r_{\langle i \rangle}^{*}$, (\ref{a1}) and (\ref{a2}) are equivalent to the OCBA formula in \cite{chen2000simulation}, which is derived under a static optimization framework. Many existing sequential sampling allocation procedures such as KG and expected improvement (EI) (see \cite{ryzhov2015asymptotic}) cannot achieve the asymptotically optimal sampling ratio. Another AOAP that sequentially achieves the OCBA formula can be found in \cite{peng2015asymptotic}, but was derived from a more tractable surrogate optimization formulation.  
The novelty of AOAP (\ref{map}) is that it is derived from an ADP framework and due to its analytical form given in (\ref{map}), it is computationally more efficient than existing sequential sampling allocation procedures such as KG, EI, and the AOAP in \cite{peng2015asymptotic}. 
\subsection{Generalizations in ADP}\label{gadp}
We can further extend the certainty equivalent approximating scheme (\ref{ax}) to a VFA that looks $b$ steps ahead into the future by the following recursion: for $1\leq\ell\leq b-1$, 
\begin{align*}
&\bar{V}_{t+b-1}^{(b)}(\mathcal{E}_{t};i_1,\ldots,i_{b})\\
&\ed\widetilde{V}_{t+b-1}\left(\mathcal{E}_{t},\left.\mathbb{E}\left[X_{i_1,t+1}\right|\mathcal{E}_{t}\right], \ldots, \left.\mathbb{E}\left[X_{i_b,t+b}\right|\mathcal{E}_{t}\right]\right),\\
&\bar{V}_{t+\ell-1}^{(b)}\left(\mathcal{E}_{t};i_1,\ldots,i_{\ell}\right)=\max_{i_{\ell+1}=1,\ldots,k}\bar{V}_{t+\ell}^{(b)}\left(\mathcal{E}_{t},i_1,\ldots,i_{\ell+1}\right),
\end{align*}
and the sampling allocation policy looking $b$ steps ahead is given by 
\begin{equation*}\label{map2}
\bar{A}_{t+1}(\mathcal{E}_{t})=\arg\max_{i_1=1,\ldots,k}\bar{V}_{t}^{(b)}(\mathcal{E}_{t};i_1)~.\end{equation*}

For value function (\ref{vf}), 
 \begin{align*}
& \mathbb{E}[V_P(\theta;\langle 1 \rangle_t)| \mathcal{E}_t]=P\left( \mu_{\langle 1 \rangle_t}>\mu_{\langle i \rangle_t},~i\neq 1|\mathcal{E}_t\right)\\
 = &P\left( \frac{\mu_{\langle 1 \rangle_t}-\mu_{\langle i \rangle_t}-\left( \mu_{\langle 1 \rangle_t}^{(t)}-\mu_{\langle i \rangle_t}^{(t)} \right)}{\sqrt{\left(\sigma_{\langle 1 \rangle_t}^{(t)}\right)^2+\left(\sigma_{\langle i \rangle_t}^{(t)}\right)^2}}>\right.\\
  &\left.\left.\qquad\qquad-\frac{\mu_{\langle 1 \rangle_t}^{(t)}-\mu_{\langle i \rangle_t}^{(t)}}{\sqrt{\left(\sigma_{\langle 1 \rangle_t}^{(t)}\right)^2+\left(\sigma_{\langle i \rangle_t}^{(t)}\right)^2}},~i\neq 1\right|\mathcal{E}_t\right)\\
  = &P\left( \widetilde{Z}_i> -d_i(\mathcal{E}_t),~i\neq 1|\mathcal{E}_t\right), \end{align*}
  where $(\widetilde{Z}_2,\ldots,\widetilde{Z}_k)$ follows a multivariate normal distribution with mean zero, variances all ones, and correlations given by $i,j\neq 1$, $i\neq j$, 
  $$\rho_{i,j}(\mathcal{E}_t)=\frac{\left(\sigma_{\langle 1 \rangle_t}^{(t)}\right)^2}{\sqrt{\left(\sigma_{\langle 1 \rangle_t}^{(t)}\right)^2+\left(\sigma_{\langle i \rangle_t}^{(t)}\right)^2}\sqrt{\left(\sigma_{\langle 1 \rangle_t}^{(t)}\right)^2+\left(\sigma_{\langle j\rangle_t}^{(t)}\right)^2}},$$
  which are called the induced correlations in \cite{peng2016gradient}, because they are induced by the variance of $\mu_{\langle 1 \rangle_t}$. From the above rewriting, we know that the value function is a function of $d_i$, $i=2,\ldots,k$, and $\rho_{i,j}$, $i,j\neq 1$, $i\neq j$. 
  
  For AOAP (\ref{map}), we can see the induced correlations are ignored in VFA (\ref{ax}). From Theorem \ref{thm3}, we know that AOAP (\ref{map}) sequentially achieves the asymptotically optimal sampling rate of the PCS, which implies the induced correlations are not significant factors for the  value function (\ref{ax}) when the simulation budget is large enough. However, \cite{peng2015nonmonotone} and \cite{peng2016gradient} showed that the induced correlations are significant factors in a low-confidence scenario that are qualitatively described by three characteristics: the differences between means of competing alternatives are small, the variances are large, and the simulation budget is small.  
  \cite{peng2016gradient} provided an efficient sequential algorithm using an analytical approximation of value function (\ref{vf}) for the low-confidence scenarios. Here, we provide an alternative algorithm using a two-factor VFA and a gradient-based Monte Carlo learning (G-MCL) scheme to fit the weights of two factors.
  
  We first establish the general results of the G-MCL scheme for optimally approximating value function (\ref{vf}) by a parametric family of VFAs (\ref{nn}).  Assuming that for $w\in W\ed\{ w\in\mathbb{R}^{\tau}:~ 0\leq w_j\leq \bar{w},~j=1,\ldots,\tau \}$, where  $\bar{w}$ is an arbitrarily large constant, 
        $$\mathbb{E}\left[\bar{V}^2(\mathcal{E}_t;w) \right]<\infty,$$
  an optimal VFA in (\ref{nn}) can be defined by $\bar{V}(\mathcal{E}_t;w^{*})$, where $w^{*}$ is the solution of the following  least-squares problem (LSP):
    \begin{align}\label{op1}
   w^{*}=\arg\min_{w\in W} \mathbb{E}\left[ \left(\bar{V}(\mathcal{E}_t;w)-\mathbb{E}[V_P(\theta;\langle 1 \rangle_t)| \mathcal{E}_t]\right)^2\right],
    \end{align}
   with  $\mathcal{E}_t$ being the information set generated by a sampling procedure independent of $w$, e.g., equal allocation.
        With the optimal VFA $\widetilde{V}(\mathcal{E}_t)=\bar{V}(\mathcal{E}_t;w^{*})$, we can use the certainty equivalence approximation to derive a one-step look ahead policy or multi-step look ahead extension. 
      
      Notice that the objective function in (\ref{op1}) involves an expectation of a nonlinear function of a conditional expectation, which is generally computationally intensive to estimate by Monte Carlo simulation. However, for LSP (\ref{op1}), the optimal solution can be efficiently found by the following stochastic approximation (SA) search algorithm (see \cite{yin2003stochastic}) with a single-run gradient estimate (see \cite{fu2015gradient}) as an input in each iteration of the SA algorithm: 
     \begin{align}\label{sa}w^{(l+1)}=\Pi_{W}\left( w^{(l)}+\lambda_l D(\mathcal{E}_t^{(l)};w^{(l)})\right) ,\end{align}
       where 
       \begin{align*}
       &D(\mathcal{E}_t^{(l)};w^{(l)})\\
       &\ed\left(\bar{V}(\mathcal{E}_t^{(l)};w^{(l)})-{\bf1}\{ \widehat{\mathcal{S}}(\mathcal{E}_t^{(l)})=\langle 1 \rangle \}\right)\nabla_w \bar{V}(\mathcal{E}_t^{(l)};w)|_{w=w^{(l)}},  \end{align*}
      $\mathcal{E}_t^{(l)}$ is the $l$-th independent realization of the information set $\mathcal{E}_t$, and $\Pi_{W}(\cdot)$ is an operator that projects the argument onto the compact feasible set $W$. Define 
       $$J(w)\ed\mathbb{E}\left[\left(\bar{V}(\mathcal{E}_t;w)-{\bf1}\{ \widehat{\mathcal{S}}(\mathcal{E}_t)=\langle 1 \rangle \}\right)^2\right]~.$$
       To justify the convergence of SA (\ref{sa}), a set of regularity conditions can be introduced as follows.
       \begin{itemize}
       \item[(i)] $J(w)$ is convex on $W$.
       \item[(ii)] For any $w\in W$, the gradient estimator $ D(\mathcal{E}_t;w)$ is unbiased and has bounded second moment, i.e., $$\nabla_w J(w)/2=\mathbb{E}\left[ D(\mathcal{E}_t;w) \right],\qquad  \mathbb{E}\left[ D^2(\mathcal{E}_t;w)\right]<\infty~.$$
       \item [(iii)] The step size sequence $\{\lambda_l\}$ satisfies the condition: $\sum_{l=1}^{\infty} \lambda_l=\infty$ and $\sum_{l=1}^{\infty} \lambda_l^2<\infty$.
       \end{itemize}
  
  \begin{theorem}\label{thm4}
  Under conditions (i) -- (iii),  $$\lim_{l\to\infty} w^{(l)}=w^{*}\quad a.s.$$ 
  \end{theorem}  \begin{IEEEproof}  The objective function of LSP (\ref{op1}) is 
      \begin{align*}
     &\mathbb{E}\left[ \left(\bar{V}(\mathcal{E}_t;w)-\mathbb{E}[V_P(\theta;\langle 1 \rangle_t)| \mathcal{E}_t]\right)^2\right]\\
     =& \mathbb{E}\left[ \bar{V}^2(\mathcal{E}_t;w)-2\bar{V}(\mathcal{E}_t;w)\mathbb{E}[V_P(\theta;\langle 1 \rangle_t)| \mathcal{E}_t]\right.\\
     &\left.\qquad\qquad\qquad\qquad\qquad\qquad\qquad+\mathbb{E}^2[V_P(\theta;\langle 1 \rangle_t)| \mathcal{E}_t]\right]~.
      \end{align*}
      Since the last term is a constant independent of $w$, the solution of  LSP (\ref{op1}) is the same as the solution of the following optimization problem:
        \begin{align*}
       w^{*}=\arg\min_{w\in W}\mathbb{E}&\left[ \bar{V}^2(\mathcal{E}_t;w)-2\bar{V}(\mathcal{E}_t;w)\mathbb{E}[V_P(\theta;\langle 1 \rangle_t)| \mathcal{E}_t]\right.\\
       &\left.\qquad\qquad\qquad\qquad\qquad+\mathbb{E}[V_P(\theta;\langle 1 \rangle_t)| \mathcal{E}_t]\right]~.
        \end{align*}
        The objective function of the above optimization problem can be rewritten as 
       \begin{align*}
       &\mathbb{E}\left[ \bar{V}^2(\mathcal{E}_t;w)-2\bar{V}(\mathcal{E}_t;w) \mathbb{E}[V_P(\theta;\langle 1 \rangle_t)| \mathcal{E}_t]+\mathbb{E}[V_P(\theta;\langle 1 \rangle_t)| \mathcal{E}_t]\right]\\
       =&\mathbb{E}\left[ \bar{V}^2(\mathcal{E}_t;w)-2\bar{V}(\mathcal{E}_t;w)\mathbb{E}[{\bf1}\{ \widehat{\mathcal{S}}(\mathcal{E}_t)=\langle 1 \rangle \}| \mathcal{E}_t]\right.\\
       &\qquad\qquad\qquad\qquad\qquad\qquad\qquad\left.+\mathbb{E}[{\bf1}\{ \widehat{\mathcal{S}}(\mathcal{E}_t)=\langle 1 \rangle \}| \mathcal{E}_t]\right]\\
       =&\mathbb{E}\left[ \mathbb{E}\left.\left[\left(\bar{V}(\mathcal{E}_t;w)-{\bf1}\{ \widehat{\mathcal{S}}(\mathcal{E}_t)=\langle 1 \rangle \}\right)^2\right| \mathcal{E}_t\right]\right]\\
       &=\mathbb{E}\left[\left(\bar{V}(\mathcal{E}_t;w)-{\bf1}\{ \widehat{\mathcal{S}}(\mathcal{E}_t)=\langle 1 \rangle\}\right)^2\right],
         \end{align*}  
         where the first equality is by definition, the second equality is because $\bar{V}(\mathcal{E}_t;w)$ is $\mathcal{E}_t$-measurable, and the third equality is due to the law of total expectation. Therefore, $$w^{*}=\arg\min_{w\in W} J(w)~.$$         
         With conditions (i)-(iii), the conclusion of the theorem can be proved using standard convergence results of SA (see \cite{yin2003stochastic}).
 \end{IEEEproof}
  \noindent\textbf{Remark.} If $J(w)$ is not convex, SA  (\ref{sa}) might converge to a local minimum. 
  For a linear VFA, we can prove convexity, thus guaranteeing the convergence to the global optimum. Specifically, for a  linear VFA, $K(z)=z$, and assuming the gradient and expectation can be interchanged, which is usually justified by the dominated convergence theorem (see \cite{rudin1987real}), 
  \begin{align*}\nabla_{w}^2 J(w)&=\mathbb{E}\left[\nabla_w^2 \left(\bar{V}(\mathcal{E}_t;w)-{\bf1}\{ \widehat{\mathcal{S}}(\mathcal{E}_t)=\langle 1 \rangle \}\right)^2 \right]\\
  &=\mathbb{E}\left[\nabla_w^2 \left(\sum_{j=1}^{\tau} w_j g_j(\mathcal{E}_t)-{\bf1}\{ \widehat{\mathcal{S}}(\mathcal{E}_t)=\langle 1 \rangle \}\right)^2 \right]\\
  &=2\mathbb{E}\left[ g'(\mathcal{E}_t) ~ g(\mathcal{E}_t) \right],\end{align*}
  where $g(\mathcal{E}_t)\ed (g_1(\mathcal{E}_t),\ldots,g_{\tau}(\mathcal{E}_t))$. It is easy to show $\nabla_{w}^2 J(w)$ is positive semi-definite and is positive definite for $w\in W$ if $g_j(\mathcal{E}_t)\geq 0$, $j=1,\ldots,\tau$, and are not identically zero. 
  
   Then, we propose the following two-factor parametric family of VFAs: 
  $$\bar{V}(\mathcal{E}_t;w)= K\left(w_1 g_1(\mathcal{E}_t)+w_2 g_2(\mathcal{E}_t)\right),$$
  where $w_1,w_2\geq 0$, $w=(w_1,w_2)$, 
  $g_1(\mathcal{E}_t)\ed d^2(\mathcal{E}_t)$, and 
  $$g_2(\mathcal{E}_t)\ed\min_{i,j\neq 1, ~i\neq j,}\rho_{i,j}^2(\mathcal{E}_t)~.$$
The feature  $g_1(\mathcal{E}_t)$ reflects the mean-variance trade-off, and $g_2(\mathcal{E}_t)$ is a feature including information on the induced correlations. From \cite{peng2015nonmonotone}, we know that  posterior IPCS (\ref{vf}) is increasing with respect to the values of both $g_1(\mathcal{E}_t)$ and $g_2(\mathcal{E}_t)$, thus  $w_1,w_2\geq 0$ is assumed. Notice that  as $t$ goes to infinity, $g_1(\mathcal{E}_t)$ goes to infinity at rate $O(t)$ while $g_2(\mathcal{E}_t)$ converges to a constant; thus the effect of the mean-variance trade-off reflected in $g_1(\mathcal{E}_t)$ will be more and more significant as $t$ grows.   If  activation function $K(\cdot)$ is monotonically increasing, the VFA of AOAP (\ref{map}) is equivalent to a special case of the AOAP based on the two-factor VFA when $w_2=0$. Besides the linear VFA,  we can also choose some nonlinear VFAs such as $K(z)=1-\exp(-z)$.
 \section{Numerical Results}\label{nex}

   We test the proposed AOAPs in Section \ref{asn} for the normal sampling distributions in the low-confidence and high-confidence scenarios. 
   A numerical example to illustrate how to calculate the optimal A\&S policy for a simple discrete sampling distribution example can be found in the online appendix (see \cite{peng2017stochastic}).
   The priors for the unknown means of the normal sampling distributions are assumed to be the normal conjugate priors introduced in Section \ref{cj}. \\
   
   \noindent\underline{Example 1: A High-Confidence Scenario.}\\
    
   In this example, we test the numerical performance of AOAP (\ref{map}) with $10$ competing alternatives in a high-confidence scenario with hyper-parameters in the prior given by $\mu_i^{(0)}=0$, and $\sigma^{(0)}_i=1$, and the true variances given by $\sigma_i=1$, $i=1,\ldots,10$. The mean and standard deviation of the true mean are
   $$\mathbb{E}\left[ \mu_i \right]=\mu_i^{(0)}=0,\quad Std\left( \mu_i \right)=\sigma^{(0)}_i=1,\qquad i=1,\ldots,10.$$
   The standard deviation of the true mean controls the dispersion of the true mean following the prior distribution. Statistically speaking, the differences in the true means of competing alternatives would be relatively large, compared with the sampling variances for simulation budget size  up to $T=400$ in this example, so this example is categorized as a high-confidence scenario. 
    The first $100$ replications are equally allocated to each alternative to estimate the sampling variances $\sigma_i^2$, $i=1,\ldots,10$. 
    
\begin{figure}[tb]
   \begin{center}
   \includegraphics[scale=0.6]{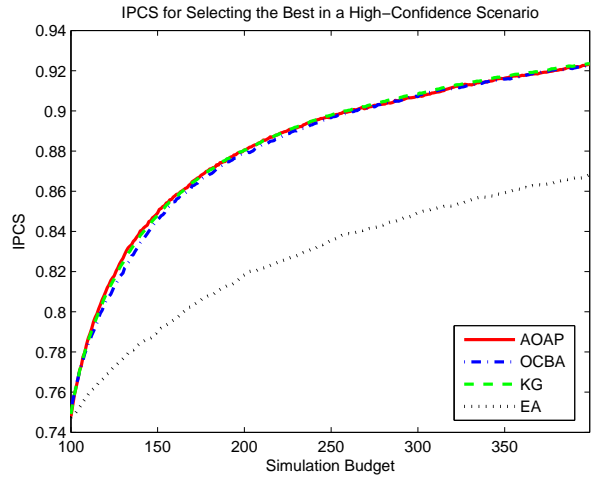}
   \caption{ The prior distribution is the normal conjugate prior, with parameters $\mu_i^{(0)}=0$, and $\sigma^{(0)}_i=1$, $i=1,\ldots,10$. The true variances are $\sigma_i=1$, $i=1,\ldots,10$. The number of initial replications is $n_0=10$ for each alternative.
   The IPCSs are estimated by $10^5$ independent macro replications.}
   \label{pic1} 
   \end{center} \end{figure}

    We compare the performance of  AOAP (\ref{map}) with the ``most starving'' sequential OCBA algorithm in \cite{chen2011stochastic},  the KG algorithm in \cite{frazier2008knowledge}, and equal allocation (EA). The selection policy is fixed as $\widehat{\mathcal{S}}(\mathcal{E}_t)=\langle 1 \rangle_t$, which selects the alternative with the largest posterior mean. 
    The performance of each sampling procedure is measured by the IPCS  following the sampling procedure to allocate a fixed amount of simulation budget, i.e., 
    $$IPCS_t\ed\mathbb{E}\left[{\bf1}\{ \widehat{\mathcal{S}}(\mathcal{E}_t)=\langle 1 \rangle \}\right]~.$$
    The IPCSs are reported in Figure \ref{pic1} as functions of the simulation budget $t$ up to $T=400$. The statistics are estimated from $10^5$ independent macro simulations. 
    
   From Figure \ref{pic1}, we can see that the three sequential policies have comparable performance that is significantly better than EA. Although OCBA and KG are derived from surrogate optimization problems, the numerical result indicates that for this high-confidence scenario, their performances are quite close to AOAP (\ref{map}). AOAP, OCBA, KG, and EA take around 1.1s, 0.8s, 7s, and 0.1s, respectively, to allocate $400$ replications.
   Numerical comparison between AOAP, OCBA, KG, and EA in the presence of correlation can be found in the online appendix (see \cite{peng2017stochastic}).\\
   
    \noindent\underline{Example 2: Two Low-Confidence Scenarios.}\\
   
     We first test the numerical performance of an AOAP with a two-factor linear VFA. In this example, there are $10$ competing alternatives in a low-confidence scenario with hyper-parameters in the prior given by $\mu_i^{(0)}=0$, $i=1,\ldots,10$,  $\sigma^{(0)}_{1}=0.02$, and $\sigma^{(0)}_i=0.01$, $i=2,\ldots,10$, and the true variances given by $\sigma_i=1$, $i=1,\ldots,10$. Statistically speaking, the differences in the true means of competing alternatives would be relatively small, compared with the sampling variances for simulation budget size up to $T=200$ in this example, so this example is categorized as a low-confidence scenario.

 We use the two-factor linear VFA proposed in Section \ref{gadp} to take the information of the induced correlation into account. The numerical results for a two-factor nonlinear VFA can be found in the online appendix. The weights of the two factors are fitted by the G-MCL scheme. In principle, the weights should be fitted for every step  up to $T=200$, because the SCP for R\&S is a nonstationary MDP. However, for computational simplicity, we only fit the weights at the final step $T$, and use the same fitted weights throughout all steps of the allocation decisions. The sampling algorithm generating $\mathcal{E}_T$ in G-MCL is EA. 
  The step size and the starting point of SA is chosen as $\lambda_l=10\times l^{-\frac{2}{3}}$ and $(w_1^{(0)},w_2^{(0)})=(1,1)$. The trajectory of SA is shown in the online appendix, and the final fitting results are $w_1^{*}\approx 0.98$ and $w_2^{*}\approx 0.42$. The rest of the sampling allocation experiment is designed the same as the last example.

      \begin{figure}[tb]
      \begin{center}
      \includegraphics[scale=0.6]{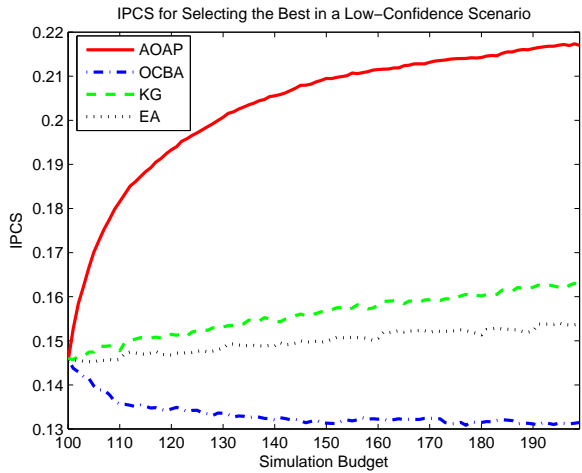}
      \caption{ The prior distribution is the normal conjugate prior, with parameters $\mu_i^{(0)}=0$, $i=1,\ldots,10$, $\sigma^{(0)}_{1}=0.02$, and $\sigma^{(0)}_i=0.01$, $i=2,\ldots,10$. The true variances are $\sigma_i=1$, $i=1,\ldots,10$. The number of initial replications is $n_0=10$ for each alternative. IPCSs are estimated by $10^5$ independent macro replications.}
      \label{pic2} 
      \end{center} \end{figure}  
      
      From Figure \ref{pic2}, we can see the IPCS of OCBA decreases as the simulation budget grows, whereas the IPCSs of KG and EA increase with the simulation budget at a slow pace, and the former has a slight edge over the latter; in contrast, the IPCS of the proposed AOAP using a two-factor VFA increases at a fast rate and  is significant larger than the IPCSs obtained by the other methods. This phenomenon is due to the fact that the information of the induced correlations, which is significant in the low-confidence scenario, is ignored or not taken fully into account by OCBA and KG. A detailed theoretical analysis on the property of the PCS and more numerical results on the performances of many existing methods in the low-confidence scenario can be found in \cite{peng2015nonmonotone}.

           \begin{figure}[tb]
             \begin{center}
             \includegraphics[scale=0.6]{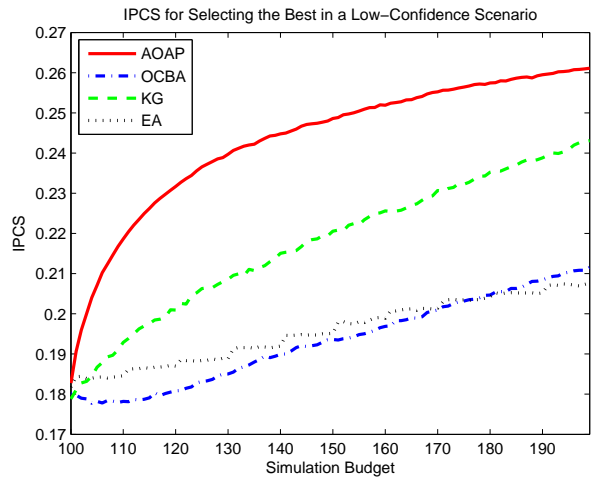}
             \caption{ The prior distribution is the normal conjugate prior, with parameters $\mu_i^{(0)}=0$, $i=1,\ldots,10$, $\sigma^{(0)}_{1}=0.08$, and  $\sigma^{(0)}_i=0.04$, $i=2,\ldots,10$. The true variances are $\sigma_i=1$, $i=1,\ldots,10$. The number of initial replications is $n_0=10$ for each alternative. IPCSs are estimated by $10^5$ independent macro replications.}
             \label{pic4} 
             \end{center} \end{figure} 
             
          We also provide a scenario that lies between Example 1 and the last example. 
     The hyper-parameters in the prior are set by $\mu_i^{(0)}=0$, $i=1,\ldots,10$,  $\sigma^{(0)}_{1}=0.08$, and $\sigma^{(0)}_i=0.04$, $i=2,\ldots,10$, and the true variance given by $\sigma_i=1$, $i=1,\ldots,10$. We can see the dispersion of the true means are larger than the last example but much smaller than Example 1. We still use an AOAP with a two-factor linear VFA  fitted by the G-MCL scheme. The initialization of the experiment is set the same as the last example. The final fitting results are $w_1^{*}\approx 0.22$ and $w_2^{*}\approx 0.53$. From Figure \ref{pic4}, we can see that the IPCS of OCBA decreases slightly at the beginning, and then quickly catches up with the IPCS of EA and surpasses the latter at the end. KG has a significant edge over OCBA and EA, but lags behind AOAP. 

   \section{Conclusion}\label{cls}
   We propose a SCP to formulate the sequential A\&S decision for the Bayesian framework of R\&S and derive the associated Bellman equation. We further analyze the optimal selection policy and the computational complexity of the optimal A\&S policy for discrete sampling and prior distributions. We especially focus on developing efficient techniques to approximate the optimal A\&S policy for the posterior IPCS of independent normal sampling distributions. An AOAP using a single feature of the posterior IPCS is proved to be asymptotically optimal. We propose a general G-MCL scheme to optimally fit the posterior IPCS by VFA. A generalized AOAP using a two-factor VFA with the G-MCL scheme can achieve a significant efficiency enhancement in the low-confidence scenario.
   
  The establishment of the Bellman equation of the SCP in our work relies on the independence of the replications. In practice, the independence assumption might not be always satisfied, e.g., $k$ machines following stationary Markov processes. How to better formulate a SCP for non-independent replications in the Bayesian framework of R\&S is an interesting theoretical question for  future research.
   
   Another direction for future research is to incorporate more features to better describe the value function of the PCS payoff under independent normal sampling distributions or even correlated normal sampling distributions.
   It would be worthwhile to consider efficient approximation schemes for other rewards such as  PCS and EOC of subset selection (see \cite{chen2008efficient} and \cite{gao2015efficient}) and the optimal quantile selection (see \cite{peng2015bayesian}). How to develop fast learning schemes for nonstationary MDPs in R\&S  deserves further study. Utilizing the cloud computing platform to solve R\&S is also a future research (see \cite{xu2015simulation}).
\bibliographystyle{apa}
\bibliography{Peng}
%\bibliographystyle{IEEEtran} 
%\bibliography{IEEEabrv,Peng}
          
%\appendices
%\section*{Appendix}

% if have a single appendix:
%\appendix[Proof of the Zonklar Equations]
% or
%\appendix  % for no appendix heading
% do not use \section anymore after \appendix, only \section*
% is possibly needed

% use appendices with more than one appendix
% then use \section to start each appendix
% you must declare a \section before using any
% \subsection or using \label (\appendices by itself
% starts a section numbered zero.)
%

% use section* for acknowledgment
\section*{Acknowledgment}

This work was supported in part by the National Science Foundation under Awards ECCS-1462409, CMMI-1462787 and CMMI-1233376, and by the China Postdoctoral Science Foundation under Grant 2015M571495.

% Can use something like this to put references on a page
% by themselves when using endfloat and the captionsoff option.
\ifCLASSOPTIONcaptionsoff
  \newpage
\fi

% trigger a \newpage just before the given reference
% number - used to balance the columns on the last page
% adjust value as needed - may need to be readjusted if
% the document is modified later
%\IEEEtriggeratref{8}
% The "triggered" command can be changed if desired:
%\IEEEtriggercmd{\enlargethispage{-5in}}

% references section

% can use a bibliography generated by BibTeX as a .bbl file
% BibTeX documentation can be easily obtained at:
% http://mirror.ctan.org/biblio/bibtex/contrib/doc/
% The IEEEtran BibTeX style support page is at:
% http://www.michaelshell.org/tex/ieeetran/bibtex/
%\bibliographystyle{IEEEtran}
% argument is your BibTeX string definitions and bibliography database(s)
%\bibliography{IEEEabrv,\ldots/bib/paper}
%
% <OR> manually copy in the resultant .bbl file
% set second argument of \begin to the number of references
% (used to reserve space for the reference number labels box)

% biography section
% 
% If you have an EPS/PDF photo (graphicx package needed) extra braces are
% needed around the contents of the optional argument to biography to prevent
% the LaTeX parser from getting confused when it sees the complicated
% \includegraphics command within an optional argument. (You could create
% your own custom macro containing the \includegraphics command to make things
% simpler here.)
%\begin{IEEEbiography}[{\includegraphics[width=1in,height=1.25in,clip,keepaspectratio]{mshell}}]{Michael Shell}
% or if you just want to reserve a space for a photo:

\begin{IEEEbiographynophoto}{Yijie Peng} 
received the B.E. degree in mathematics from Wuhan University, China and the Ph.D. degree in management science from Fudan University, China, in 2007 and 2014, respectively.
After working as a research fellow at Fudan University and George Mason University, he joined the Department of Industrial Engineering and Management at Peking University in July, 2017.
His research interests are in ranking and selection and sensitivity analysis in the field simulation optimization.  

%Dr. Peng was awarded a National Scholarship of China and a Scholarship for Cultivating Innovative Students in Key Disciplines, Project 985 of China, both in 2013. 
\end{IEEEbiographynophoto}

\begin{IEEEbiographynophoto}{Edwin K. P. Chong}
(F'04) received the B.E. degree with First Class Honors from the University of Adelaide, South Australia, in 1987; and the M.A. and Ph.D. degrees in 1989 and 1991, respectively, both from Princeton University, where he held an IBM Fellowship. He joined the School of Electrical and Computer Engineering at Purdue University in 1991, where he was named a University Faculty Scholar in 1999. Since August 2001, he has been a Professor of Electrical and Computer Engineering and Professor of Mathematics at Colorado State University.  He coauthored the best-selling book, \emph{An Introduction to Optimization} (4th Edition, Wiley-Interscience, 2013). He received the NSF CAREER Award in 1995 and the ASEE Frederick Emmons Terman Award in 1998. He was a co-recipient of the 2004 Best Paper Award for a paper in the journal \emph{Computer Networks}. In 2010, he received the IEEE Control Systems Society Distinguished Member Award.

Prof. Chong was the founding chairman of the IEEE Control Systems Society Technical Committee on Discrete Event Systems, and served as an IEEE Control Systems Society Distinguished Lecturer. He is currently a Senior Editor of the \textsc{IEEE Transactions on Automatic Control}, and has also served on the editorial boards of \emph{Computer Networks}, \emph{Journal of Control Science and Engineering}, and \emph{IEEE Expert Now}. He has served as a member of the IEEE Control Systems Society Board of Governors and as Vice President for Financial Activities until 2014. He currently serves as President.  He was the General Chair for the 2011 Joint 50th IEEE Conference on Decision and Control and European Control Conference.
\end{IEEEbiographynophoto}

\begin{IEEEbiographynophoto}{Chun-Hung Chen}
(S'91-M'94-SM'01-F'16) received his Ph.D. degree in Engineering Sciences from Harvard University in 1994. He is a Professor of Systems Engineering and Operations Research at George Mason University. Dr. Chen received “National Thousand Talents" Award from the central government of China in 2011, the Best Automation Paper Award from the 2003 IEEE International Conference on Robotics and Automation, and 1994 Eliahu I. Jury Award from Harvard University. He has served as a Department Editor for IIE Transactions, Department Editor for Asia-Pacific Journal of Operational Research, Associate Editor for IEEE Transactions on Automation Science and Engineering, Associate Editor for IEEE Transactions on Automatic Control, Area Editor for Journal of Simulation Modeling Practice and Theory, Advisory Editor for International Journal of Simulation and Process Modeling, and Advisory Editor for Journal of Traffic and Transportation Engineering. Dr. Chen is the author of two books, including a best seller: “Stochastic Simulation Optimization: An Optimal Computing Budget Allocation”.
\end{IEEEbiographynophoto}

% if you will not have a photo at all:

% insert where needed to balance the two columns on the last page with
% biographies
%\newpage

\begin{IEEEbiographynophoto}{Michael C. Fu}
(S'89- M'89-SM'06-F'08) received degrees in mathematics and electrical engineering and computer science from MIT, Cambridge, MA, in 1985 and the Ph.D. degree in applied math from Harvard University in 1989.
Since 1989, he has been with the University of Maryland, College Park. He has also served as the Operations Research Program Director at the National Science Foundation and is a Fellow of the Institute for Operations Research and the Management Sciences.
\end{IEEEbiographynophoto}

% You can push biographies down or up by placing
% a \vfill before or after them. The appropriate
% use of \vfill depends on what kind of text is
% on the last page and whether or not the columns
% are being equalized.

%\vfill

% Can be used to pull up biographies so that the bottom of the last one
% is flush with the other column.
%\enlargethispage{-5in}

% that's all folks
\end{document}